\documentclass[letterpaper]{article} 
\usepackage[preprint]{aaai24}
\usepackage{times}  
\usepackage{helvet}  
\usepackage{courier}  
\usepackage[hyphens]{url}  
\usepackage{graphicx} 
\usepackage{natbib}  
\usepackage{caption} 
\usepackage{algorithm}
\usepackage{algorithmic}

\usepackage{newfloat}
\usepackage{listings}
\DeclareCaptionStyle{ruled}{labelfont=normalfont,labelsep=colon,strut=off} 
\floatstyle{ruled}
\newfloat{listing}{tb}{lst}{}
\floatname{listing}{Listing}
\title{A Unified View of Abstract Visual Reasoning Problems}
\author{%
  Mikołaj Małkiński\textsuperscript{\rm 1},
  Jacek Mańdziuk\textsuperscript{\rm 1,\rm 2}
}
\affiliations{%
    \textsuperscript{\rm 1}Faculty of Mathematics and Information Science, Warsaw University of Technology, Warsaw, Poland\\
    \textsuperscript{\rm 2}Faculty of Computer Science, AGH University of Krakow, Krakow, Poland\\
    mikolaj.malkinski.dokt@pw.edu.pl, mandziuk@mini.pw.edu.pl
}

\usepackage[table,x11names,dvipsnames]{xcolor}

\usepackage{amsmath}
\usepackage{amssymb}
\usepackage{mathtools}
\usepackage{amsthm}
\usepackage{bm}
\usepackage{amsfonts}  
\usepackage{subcaption}
\usepackage{multirow}  
\usepackage{makecell}  
\usepackage{textcomp}  
\usepackage{numprint}  
\usepackage{url}            
\usepackage{booktabs}       
\npthousandsep{,}
\usepackage{soul}

\usepackage{algorithm}  
\usepackage{algorithmic}  
\usepackage{varwidth}

\usepackage{nicefrac}       

\usepackage{etoolbox}
\usepackage{tikz}
\usetikzlibrary{calc,shapes.geometric,trees,positioning,arrows,arrows.meta,backgrounds,fit,decorations.pathreplacing,decorations.markings,calligraphy,matrix}

\makeatletter
\DeclareRobustCommand{\rvdots}{%
  \vbox{
    \baselineskip4\p@\lineskiplimit\z@
    \kern-\p@
    \hbox{.}\hbox{.}\hbox{.}
  }%
}
\makeatother

\newcommand{\unit}{0.4cm}

\newcommand*{\numans}[1]{\parbox[c]{0.04\textwidth}{\centering #1}}%

\definecolor{myred}{HTML}{FC8D59}
\definecolor{myyellow}{HTML}{FFFFBF}
\definecolor{myblue}{HTML}{91BFDB}

\definecolor{c0}{HTML}{9b9b7a}
\definecolor{c1}{HTML}{baa587}
\definecolor{c2}{HTML}{e8ac65}
\definecolor{c3}{HTML}{d08c60}

\newcommand{\CellColor}{SkyBlue!10}
\newcommand{\cl}{\cellcolor{\CellColor}}

\begin{document}


\maketitle

\begin{abstract}
The field of Abstract Visual Reasoning (AVR) encompasses a wide range of problems, many of which are inspired  by human IQ tests.
The variety of AVR tasks has resulted in state-of-the-art AVR methods being task-specific approaches.
Furthermore, contemporary methods consider each AVR problem instance not as a whole, but in the form of a set of individual panels with particular locations and roles (context vs. answer panels) pre-assigned according to the task-specific arrangements.
While these highly specialized approaches have recently led to significant progress in solving particular AVR tasks, considering each task in isolation hinders the development of universal learning systems in this domain.
In this paper, we introduce a unified view of AVR tasks, where each problem instance is rendered as a single image, with no a priori assumptions about the number of panels, their location, or role.
The main advantage of the proposed unified view is the ability to develop universal learning models applicable to various AVR tasks.
What is more, the proposed approach inherently facilitates transfer learning in the AVR domain, as various types of problems share a common representation. 
The experiments conducted on four AVR datasets with Raven's Progressive Matrices and Visual Analogy Problems, and one real-world visual analogy dataset show that the proposed unified representation of AVR tasks poses a challenge to state-of-the-art Deep Learning (DL) AVR models and, more broadly, contemporary DL image recognition methods.
In order to address this challenge, we introduce the \textit{Unified Model for Abstract Visual Reasoning (UMAVR)} capable of dealing with various types of AVR problems in a unified manner.
UMAVR outperforms existing AVR methods in selected single-task learning experiments, and demonstrates effective knowledge reuse in transfer learning and curriculum learning setups.
\end{abstract}

\definecolor{rpm}{HTML}{ffc107}
\definecolor{vap}{HTML}{26a69a}

\tikzstyle{emptylayer} = [rectangle, inner sep=0, minimum height=1.2*\unit, minimum width=4*\unit]
\tikzstyle{layer} = [emptylayer, rounded corners, text centered, draw=black]
\tikzstyle{background} = [draw, fill=black!5, rounded corners=5pt, densely dashed, inner sep=0]
\tikzstyle{panel_background} = [draw, fill=black!5, rounded corners=3pt, densely dashed, inner sep=0.15*\unit]
\tikzstyle{panel} = [thick, rectangle, draw=black, inner sep=0]
\tikzstyle{label} = [rectangle, inner sep=0]

\tikzstyle{embedding} = [circle, text centered, draw=black, inner sep=0, minimum height=1.2*\unit, minimum width=1.2*\unit]

\tikzstyle{arrow} = [->,>={Latex[scale=1]}]

\newcommand{\rpm}[1]{\includegraphics[width=0.6cm]{images/rpm/context_#1}}
\newcommand{\rpmans}[1]{\includegraphics[width=0.6cm]{images/rpm/answer_#1}}
\newcommand{\vap}[1]{\includegraphics[width=0.6cm]{images/vap/context_#1}}
\newcommand{\vapans}[1]{\includegraphics[width=0.6cm]{images/vap/answer_#1}}
\newcommand{\bvap}[1]{\includegraphics[width=0.8cm]{images/vap/context_#1}}
\newcommand{\bvapans}[1]{\includegraphics[width=0.8cm]{images/vap/answer_#1}}

\begin{figure*}[t]
    \centering
    \resizebox{0.84\textwidth}{!}{
    \begin{tikzpicture}
        \begin{scope}[node distance=7pt]
            \node (rp1) [panel] {\rpm{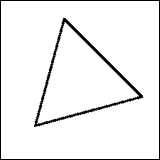}};
            \node (rp2) [panel, below=of rp1] {\rpm{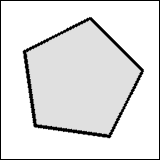}};
            \node (rpi) [text centered, inner sep=0, below=of rp2, minimum height=0.2cm, minimum width=0.2cm] {\rvdots};
            \node (rpn) [panel, below=of rpi] {\rpmans{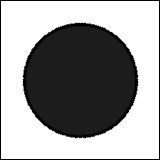}};
        \end{scope}
        \begin{scope}[on background layer]
            \node (rp1bg) [fit={(rp1)}, panel_background, fill=rpm!35] {};
            \node (rp2bg) [fit={(rp2)}, panel_background, fill=rpm!35] {};
            \node (rpnbg) [fit={(rpn)}, panel_background, fill=rpm!35] {};
        \end{scope}
        \node (rdt) [inner sep=0pt] at ($ (rp1.north) + (0, 0.9*\unit) $) {\small $\{\bm{x}^{\text{RPM}}_j\}$};

        \begin{scope}[node distance=7pt]
            \node (vp1) [panel, right=of rp1, xshift=0.9*\unit] {\vap{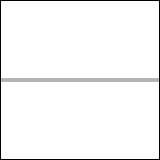}};
            \node (vp2) [panel, below=of vp1] {\vap{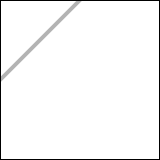}};
            \node (vpi) [text centered, inner sep=0, below=of vp2, minimum height=0.2cm, minimum width=0.2cm] {\rvdots};
            \node (vpn) [panel, below=of vpi] {\vapans{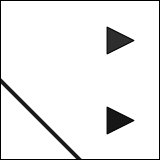}};
        \end{scope}
        \begin{scope}[on background layer]
            \node (vp1bg) [fit={(vp1)}, panel_background, fill=vap!35] {};
            \node (vp2bg) [fit={(vp2)}, panel_background, fill=vap!35] {};
            \node (vpnbg) [fit={(vpn)}, panel_background, fill=vap!35] {};
        \end{scope}
        \node (vdt) [inner sep=0pt] at ($ (vp1.north) + (0, 0.9*\unit) $) {\small $\{\bm{x}^{\text{VAP}}_j\}$};

        \begin{scope}[node distance=1pt]
            \node (r1) [panel, right=of vp1, xshift=2.4*\unit] {\rpm{0}};
            \node (r2) [panel, right=of r1] {\rpm{1}};
            \node (r3) [panel, right=of r2] {\rpm{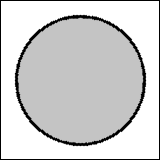}};
            \node (r4) [panel, below=of r1] {\rpm{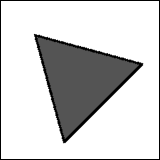}};
            \node (r5) [panel, below=of r2] {\rpm{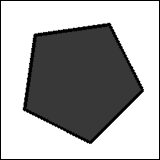}};
            \node (r6) [panel, below=of r3] {\rpm{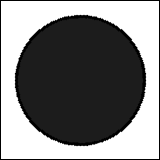}};
            \node (r7) [panel, below=of r4] {\rpm{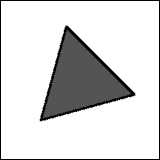}};
            \node (r8) [panel, below=of r5] {\rpm{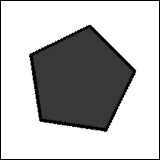}};
            \node (r9) [panel, below=of r6] {\rpm{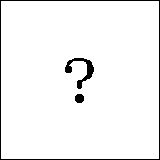}};
            \node (ra1) [panel, below=of r7, xshift=11pt, yshift=-8pt] {\rpmans{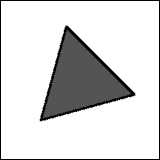}};
            \node (ra1t) [label, above=of ra1] {\footnotesize A};
            \node (ra2) [panel, right=of ra1] {\rpmans{7}};
            \node (ra2t) [label, above=of ra2] {\footnotesize B};
        \end{scope}
        \begin{scope}[on background layer]
            \node (rbg) [fit={(r1.north west) (r3.east) (ra1.south)}, background, fill=rpm!35, inner sep=0.25*\unit] {};
        \end{scope}
        \node (rut) [inner sep=0pt] at ($ (rbg.north) + (0, 0.6*\unit) $) {\small $\bm{\chi}^{\text{RPM}}$};

        \begin{scope}[node distance=1pt]
            \node (v1) [panel, right=of r3, xshift=1.3*\unit, yshift=-1.5*\unit] {\vap{0}};
            \node (v2) [panel, right=of v1] {\vap{1}};
            \node (v3) [panel, right=of v2] {\vap{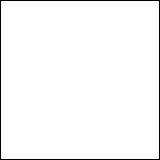}};
            \node (v4) [panel, below=of v1] {\vap{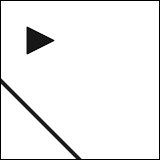}};
            \node (v5) [panel, below=of v2] {\vap{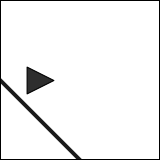}};
            \node (v6) [panel, below=of v3] {\vap{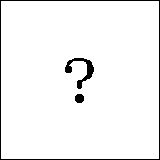}};

            \node (va1) [panel, below=of v4, xshift=11pt, yshift=-8pt] {\vapans{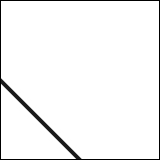}};
            \node (va1t) [label, above=of va1] {\footnotesize A};
            \node (va2) [panel, right=of va1] {\vapans{1}};
            \node (va2t) [label, above=of va2] {\footnotesize B};
        \end{scope}
        \begin{scope}[on background layer]
            \node (vbg) [fit={(v1.north west) (va1.south) (v3.east)}, background, fill=vap!35, inner sep=0.25*\unit] {};
        \end{scope}
        \node (vut) [inner sep=0pt] at ($ (vbg.north) + (0, 0.6*\unit) $) {\small $\bm{\chi}^{\text{VAP}}$};

        \begin{scope}[node distance=0.5*\unit]
            \node (re) [layer, fill=rpm!50, below=of rpn.east, xshift=0.3*\unit, yshift=-2.5*\unit] {Model};
            \node (ry) [embedding, fill=rpm!80, below=of re] {$\widehat{y}$};

            \node (ve) [layer, fill=vap!50, right=of re, xshift=0.5*\unit] {Model};
            \node (vy) [embedding, fill=vap!80, below=of ve] {$\widehat{y}$};

            \node (ae) [layer, shade, left color=rpm!50, right color=vap!50, right=of ve, xshift=1*\unit, minimum width=8*\unit] {Unified Model};
            \node (ay) [embedding, shade, left color=rpm!80, right color=vap!80, below=of ae] {$\widehat{y}$};
        \end{scope}

        \begin{scope}[on background layer]
            \node (rmbg) [fit={(re) (ry)}, background, inner sep=0.25*\unit] {};
        \end{scope}
        \begin{scope}[on background layer]
            \node (vmbg) [fit={(ve) (vy)}, background, inner sep=0.25*\unit] {};
        \end{scope}
        \node [align=center] at ($ (rmbg.south)!0.5!(vmbg.south) - (0, 1.25*\unit) $) {\footnotesize a) Disjoint view (Section~\ref{sec:disjoint-perspective})\\\footnotesize Single-task learning (STL)};

        \begin{scope}[on background layer]
            \node (ambg) [fit={(ae) (ay)}, background, inner sep=0.25*\unit] {};
        \end{scope}
        \node (unified_caption) [align=center] at ($ (ambg.south) - (0, 1.25*\unit) $) {\footnotesize b) Unified view (Section~\ref{sec:unified-perspective})\\\footnotesize Transfer learning (TL)};

        \begin{scope}[decoration={markings, mark=at position 0.5 with {\arrow[scale=1.5,>=stealth]{>}}}] 
            \draw[postaction={decorate}, densely dotted] (rpnbg.south) to[out=270, in=90] (re.north);
            \draw[postaction={decorate}, densely dotted] (vpnbg.south) to[out=270, in=90] (ve.north);;
            \draw[postaction={decorate}, solid] (rbg.south) to[out=270, in=120] (ae.north);
            \draw[postaction={decorate}, solid] (vbg.south) to[out=270, in=90] (ae.north);
        \end{scope}
        \draw (re.south) -- (ry.north);
        \draw (ve.south) -- (vy.north);
        \draw (ae.south) -- (ay.north);

        \node (sep_center) at ($ (ve.east)!0.5!(ae.west) $) {};
        \node (sep_north) at (sep_center |- rmbg.north) {};
        \node (sep_south) at (sep_center |- unified_caption.south) {};
        \draw (sep_north.north) -- (sep_south);

        \draw [-{Stealth}] (rdt.north east) to[out=30, in=180] ($ (rdt.east)!0.5!(rut.west) + (0, 15pt) $) node (renderrpm) [yshift=6pt] {\footnotesize$\mathcal{R(\cdot)}$} to[out=0, in=150] (rut.north west);
        \draw [-{Stealth}] (vdt.north east) to[out=30, in=180] ($ (vdt.east)!0.5!(vut.west) + (0, 23pt) $) node [yshift=6pt] {\footnotesize$\mathcal{R(\cdot)}$} to[out=0, in=135] (vut.north west);

        
        \begin{scope}[node distance=1pt]
            \node (v11) [panel, right=of vbg, xshift=3*\unit, yshift=3*\unit] {\bvap{0}};
            \node (v12) [panel, right=of v11] {\bvap{1}};
            \node (v13) [panel, right=of v12] {\bvap{2}};
            \node (v14) [panel, below=of v11] {\bvap{3}};
            \node (v15) [panel, below=of v12] {\bvap{4}};
            \node (v16) [panel, below=of v13] {\bvap{5}};
        \end{scope}
        \begin{scope}[on background layer]
            \node (v1bg) [fit={(v11.north west) (v16.south east)}, background, rounded corners=1pt, fill=vap!35, inner sep=0.05*\unit] {};
        \end{scope}
        \node (v1t) [inner sep=0pt, align=left, anchor=west] at ($ (v1bg.north west) + (0, 0.3*\unit) $) {\scriptsize 1. Arrange context.};
        
        \begin{scope}[node distance=1pt]
            \node (v21) [panel, right=of v13, xshift=1*\unit, yshift=-0.7*\unit] {\bvapans{0}};
            \node (v21t) [label, above=of v21] {\footnotesize A};
            \node (v22) [panel, right=of v21] {\bvapans{1}};
            \node (v22t) [label, above=of v22] {\footnotesize B};
        \end{scope}
        \begin{scope}[on background layer]
            \node (v2bg) [fit={(v21.west) (v21t.north) (v22.south east)}, background, rounded corners=1pt, fill=vap!35, inner sep=0.05*\unit] {};
        \end{scope}
        \node (v2t) [inner sep=0pt, align=left, anchor=west] at ($ (v2bg.north west) + (0, 0.3*\unit) $) {\scriptsize 2. Arrange answers.};

        \node (ralg_x) at ($ (v1bg.west)!0.5!(v2bg.east) $) {};
        \node [yshift=-1*\unit] at (ralg_x |- renderrpm) {\footnotesize $\mathcal{R} (\{\bm{x}_j^{\text{VAP}}\}_{j=1}^{P_{\text{VAP}}} \mid \mathcal{S}_{\text{VAP}}) = \bm{\chi}^{\text{VAP}}$};

        \begin{scope}[node distance=1pt]
            \node (v31) [panel, below=of v14, yshift=-1*\unit] {\bvap{0}};
            \node (v32) [panel, right=of v31] {\bvap{1}};
            \node (v33) [panel, right=of v32] {\bvap{2}};
            \node (v34) [panel, below=of v31] {\bvap{3}};
            \node (v35) [panel, below=of v32] {\bvap{4}};
            \node (v36) [panel, below=of v33] {\bvap{5}};

            \node (v3a1) [panel, below=of v34, xshift=11pt, yshift=-8pt] {\bvapans{0}};
            \node (v3a1t) [label, above=of v3a1] {\footnotesize A};
            \node (v3a2) [panel, right=of v3a1] {\bvapans{1}};
            \node (v3a2t) [label, above=of v3a2] {\footnotesize B};
        \end{scope}
        \begin{scope}[on background layer]
            \node (v3bg) [fit={(v31.north west) (v3a1.south) (v33.east)}, background, rounded corners=1pt, fill=vap!35, inner sep=0.05*\unit] {};
        \end{scope}
        \node (v3t) [inner sep=0pt, align=left, anchor=west] at ($ (v3bg.north west) + (0, 0.3*\unit) $) {\scriptsize 3. Combine.};
        
        \begin{scope}[node distance=1pt]
            \node (v41) [panel, right=of v33, xshift=1*\unit, yshift=0.05*\unit] {\vap{0}};
            \node (v42) [panel, right=of v41] {\vap{1}};
            \node (v43) [panel, right=of v42] {\vap{2}};
            \node (v44) [panel, below=of v41] {\vap{3}};
            \node (v45) [panel, below=of v42] {\vap{4}};
            \node (v46) [panel, below=of v43] {\vap{5}};

            \node (v4a1) [panel, below=of v44, xshift=11pt, yshift=-8pt] {\vapans{0}};
            \node (v4a1t) [label, above=of v4a1] {\footnotesize A};
            \node (v4a2) [panel, right=of v4a1] {\vapans{1}};
            \node (v4a2t) [label, above=of v4a2] {\footnotesize B};
        \end{scope}
        \begin{scope}[on background layer]
            \node (v4bg) [fit={(v41.north west) (v4a1.south) (v43.east)}, background, fill=vap!35, inner sep=0.25*\unit] {};
        \end{scope}
        \node (v4t) [inner sep=0pt, align=left, anchor=west] at ($ (v4bg.north west) + (0, 0.3*\unit) $) {\scriptsize 4. Resize and center.};

        \node (alg_caption) [align=center] at (ralg_x |- unified_caption) {\footnotesize c) Rendering algorithm\\\footnotesize(Section~\ref{sec:unified-perspective})};
    \end{tikzpicture}
    }
    \caption{
    \textbf{Disjoint vs. unified perspective.}
    Contemporary literature considers each AVR problem instance as a set of separate images (a), which leads to task-specific methods with limited applicability to other, even similar, tasks.
    In contrast, we propose the unified view (b) in which the problem instance is rendered as a single image (c).
    This viewpoint facilitates the development of general AVR solving models that are inherently capable of incorporating advances from broader CV research.
    }
    \label{fig:intro}
\end{figure*}
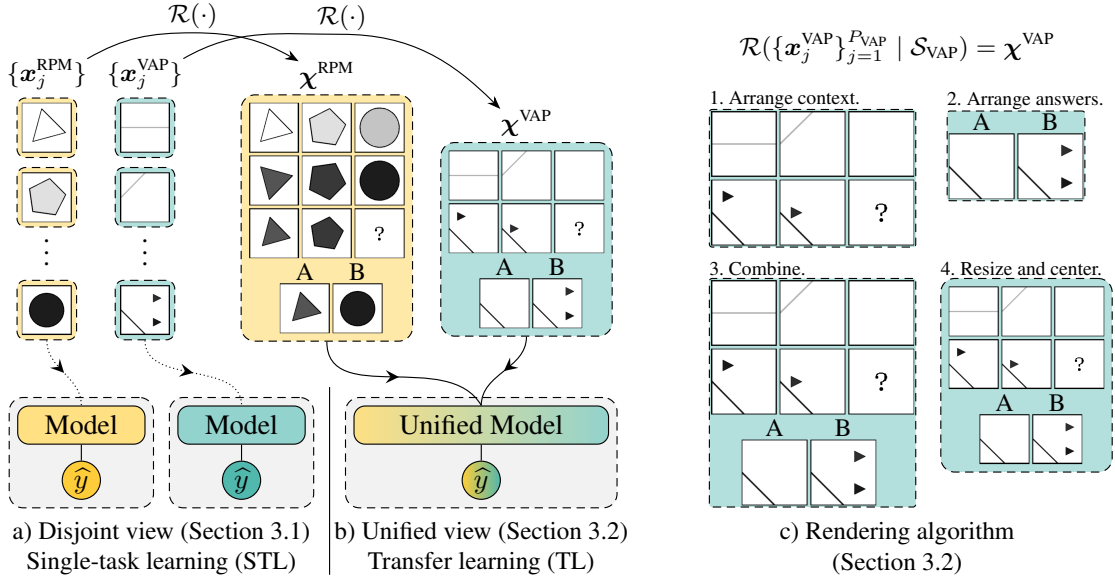

\section{Introduction}\label{sec:introduction}

Recent years have brought dynamic progress in the application of neural networks to computer vision (CV) problems.
This increasing interest has led to the development of a broad family of effective vision models based on convolutional networks~\cite{hinton2012imagenet,he2016deep,liu2022convnet}, transformers~\cite{dosovitskiy2021an,liu2021swin}, and multi-layer perceptrons (MLPs)~\cite{tolstikhin2021mlp,hou2022vision}, that often generalize well to other vision tasks.
The universality of these models should be primarily attributed to the simplicity and wide applicability of the typical problem representation in the CV domain, in the form of a single image.

Abstract Visual Reasoning (AVR) is one of the CV subdomains gaining momentum in recent years.
AVR encompasses problems that resemble tests used for measuring human abstract intelligence (IQ).
A typical example are Raven's Progressive Matrices (RPMs)~\cite{raven1936mental,raven1998raven} that consist of simple 2D shapes (e.g., circles, hexagons, triangles) characterized by several attributes (e.g., rotation, colour, size).
In most cases, an RPM problem instance is in the form of a $3 \times 3$ grid of panels, with the bottom-right panel missing.
The test-taker has to complete the matrix by selecting one of the provided answer panels.
In RPM datasets, there are usually up to 8 answer panels to choose from.
In order to select the correct one, the subject has to identify various underlying abstract rules (e.g., progression, constancy, conjunction) that govern the location and attributes of RPM shapes.
The selected answer has to conform to all these rules after being placed in the bottom-right corner of the matrix. 

RPMs are considered to be highly indicative for human intelligence~\cite{carpenter1990one}, as they allow evaluating relational and abstract reasoning skills~\cite{snow1984topography}, and test one's ability to apply previously gained knowledge in new settings (new problem instances).
Inspired by this crucial role of RPMs in designing IQ tests, recent streams of research have focused on evaluating the capacity of modern learning systems in solving RPM instances~\cite{malkinski2022deep}.
Motivated by the early successes~\cite{hoshen2017iq,mandziuk2019deepiq}, many approaches have been subsequently proposed that gradually improved the state-of-the-art (SOTA)~\cite{hu2021stratified,wu2020scattering}.

Despite impressive results, contemporary methods are built on a strict assumption that RPMs (or AVR tasks in general) are split beforehand into a set of individual matrix panels.
In stark contrast, vision datasets typically render each problem instance as a single image.
In this work, we coin these two perspectives as \emph{disjoint} and \emph{unified} representations, respectively (see Fig.~\ref{fig:intro}).
Due to inherent differences, models developed for one representation are not directly applicable to the other one.
This, in turn, limits applicability of methods developed for solving AVR tasks to other vision problems and, \textit{vice versa}, prevents evaluation of modern vision models on AVR tasks.

In addition to RPMs, other kinds of AVR benchmarks have recently been proposed~\cite{malkinski2022review}, that specifically focus on conceptual abstraction~\cite{hill2018learning}, extrapolation~\cite{webb2020learning}, or arithmetic reasoning~\cite{zhang2020machine}.
This variety of AVR problems further exacerbates the issues arising from the disjoint perspective, as AVR tasks often differ w.r.t. the number and arrangement of panels the matrix is composed of.
Consequently, a model constructed to handle tasks with a fixed number of panels arranged in a fixed configuration isn't directly capable of handling novel problem configurations, or problems with different numbers of panels.
In effect, SOTA solutions in the AVR literature are task-specific, which limits progress towards general AVR solvers.

\paragraph{Contribution.}
In this work we pose and address the challenge of building AVR models capable of solving diverse AVR problems.
To this end, we:

\noindent
(1) Formulate a \textit{unified view} of AVR tasks where a problem instance is represented as a single image (as opposed to a set of pre-defined panels), thus constituting a challenge for modern AVR/CV methods;

\noindent
(2) Evaluate different CV models: convolutional networks, transformers, and MLPs on four AVR datasets with RPMs and Visual Analogy Problems (VAPs), and one real-world visual analogy dataset, represented in the unified manner, and demonstrate their limitations in this unified problem setup;

\noindent
(3) Introduce the \textit{Unified Model for Abstract Visual Reasoning (UMAVR)}, capable of effectively dealing with the unified problem representation, and outperforming strong baselines in this arrangement;

\noindent
(4) Examine the benefits of transfer learning (TL) and curriculum learning (CL)~\cite{bengio2009curriculum} within the proposed unified AVR problem formulation. The results show that both these learning setups offer promising directions for future AVR research.

On a general note, we postulate shifting the main focus of AVR research from developing dedicated task-specific models to general AVR models, capable of solving a variety of AVR problems.

\section{Related work}\label{sec:related-work}
\paragraph{Tasks.}
Recently AVR has seen rapid expansion in terms of available benchmarks.
In particular, several RPM datasets, such as PGM~\cite{santoro2018measuring}, I-RAVEN~\cite{zhang2019raven,hu2021stratified}, or G-set~\cite{mandziuk2019deepiq,tomaszewska2022duel} have been proposed.
While the benchmarks differ in the number of rendered matrices, types and attributes of the objects, degree of compositional complexity, or the number of available answers, they all present matrices with the context panels arranged in the form of a $3\times3$ grid.
VAPs~\cite{hill2018learning} form a related challenge that focuses on conceptual abstraction, and are composed of a $2\times3$ grid of panels (see Fig.~\ref{fig:intro}b).
Other AVR problems further differ from the above two tasks in the number of context panels and their structure.
For instance, the Odd One Out tests (O3)~\cite{mandziuk2019deepiq} present images arranged in a single row, while panels in Bongard Problems~\cite{bongard1968recognition,nie2020bongard} are divided into left and right ones.
Common to all AVR problems is their input representation in Machine Learning (ML) models, as an explicit set of distinct panels.
This way of representing input data is in stark contrast to typical image recognition tasks, where each input is simply presented in the form of a single image, without further division into sub-images (though, such a division is often induced as part of a training process).
In effect, existing AVR datasets cannot be easily considered when evaluating contemporary image recognition methods and are largely omitted in multi-task vision datasets, such as VTAB~\cite{zhai2019large} or Meta-dataset~\cite{triantafillou2020meta}.
The proposed unified representation of AVR tasks enables to utilize existing AVR datasets for measuring abstract reasoning skills of current (and future) CV models.
\paragraph{AVR models.}
Due to inherent differences in the number of panels and the task structure across AVR datasets, as well as their explicit panel-based arrangement, main AVR research lines focus on designing task-specific architectures that operate on matrices pre-segmented into individual panels.
Specifically, Wild Relation Network~\cite{santoro2018measuring} employs a convolutional encoder to generate embeddings of individual panels, which are later processed by a Relation Network~\cite{santoro2017simple}.
Stratified Rule-Aware Network~\cite{hu2021stratified} considers 3 different RPM hierarchies: single images, rows/columns, and pairs of rows/columns.
For each hierarchy, a separate convolutional network is used, and a set of MLPs is employed to gradually merge the obtained embeddings.
MXGNet~\cite{wang2020abstract} separately embeds matrix panels with a convolutional encoder and utilizes a graph neural network to capture relations between image embeddings.
SCL~\cite{wu2020scattering} introduces the scattering transformation that splits panel representations, processes them in parallel with an MLP, and merges the results.
The recently proposed Slot Transformer Scoring Network (STSN)~\cite{mondal2023learning} employs Slot Attention~\cite{locatello2020object} to discover objects in matrix panels and reasons over them with a transformer~\cite{vaswani2017attention}.
SCAR~\cite{malkinski2024one} introduces the structure-aware dynamic layer that adapts its computation to the considered problem instance enabling the processing of AVR tasks with diverse structure.

The underlying assumption of the above approaches is that a problem instance is already pre-segmented into individual panels.
Consequently, direct application of these methods to other image recognition tasks (where the input is formed by a single image), or other AVR tasks that differ in the number of panels or their structure is significantly hindered.
In contrast, the proposed unified perspective facilitates and encourages the development of universal ML image recognition models that can be applied to solving diverse AVR tasks and, furthermore, other vision problems.

\section{Method}\label{sec:approach}

We start with introducing the unified perspective that can be applied to virtually all AVR tasks, and then propose a novel image recognition model, well-suited to the proposed challenge.

\subsection{Current perspective (\textit{disjoint})}\label{sec:disjoint-perspective}
In general, each AVR task $t \in \mathcal{T}$, where $\mathcal{T}$ is the family of all AVR tasks, can be defined as $t = (\{\mathcal{M}_i^t\}_{i=1}^{N_t}, \mathcal{S}_t)$, where $\{\mathcal{M}_i^t\}_{i=1}^{N_t}$ is a set of $N_t$ different matrices (problem instances) and $\mathcal{S}_t$ is a task's structure (common for all instances).
In this work, we treat each AVR dataset as a separate task.
While matrices don't repeat across tasks, some tasks may have a common structure.

For each $t$, a single matrix (problem instance) from $t$ can be defined as $\mathcal{M}_i^t = (X_i^t, y_i^t)$, where $X_i^t = \{x_{i,j}^t\}_{j=1}^{P_t}$ is composed of $P_t$ panels.
Each panel $x_{i,j}^t$ is a greyscale image with height $h$ and width $w$, i.e. $x_{i,j}^t \in [0, 1]^{h \times w}$, and $y_i^t$ is the answer to the problem.
For single-choice tasks, such as RPMs or VAPs, $y_i^t$ is an index of the correct answer.

The above definition of a problem instance is rather general and doesn't include any information about the problem's specificity.
Such problem-specific metadata is expressed by the task's structure $\mathcal{S}_t$, which defines how the images should be interpreted and arranged to form a 2D problem instance.
For example, the structure of RPMs specifies that the set of images should be split into two sets: a set of $8$ context panels arranged in a $3 \times 3$ grid, with a missing panel in the bottom-right corner, and a set of up to $8$ answer panels also arranged in a grid.

In  summary, while each AVR task $t$ is defined as: 
$t = (\{(\{\bm{x}_{i,j}^t\}_{j=1}^{P_t}, \bm{y}_i^t)\}_{i=1}^{N_t}, \mathcal{S}_t)$
its actual representation (i.e. the respective AVR dataset) is a collection of images:
$t = (\{(\{\bm{x}_{i,j}^t\}_{j=1}^{P_t}, \bm{y}_i^t)\}_{i=1}^{N_t})$
with \textit{implicitly} defined structure.
Consequently, each  AVR model $\mathcal{F}_t$ for solving matrices from $t$ embeds the problem structure directly in its architecture.
A construction of such a model using a building process $\mathcal{B}$ may be defined as a composition of functions $f$, given the task's structure $\mathcal{S}_t$:
\begin{equation}
    \mathcal{B}(f_* \circ f_\# \circ \ldots \circ f_\$ \mid \mathcal{S}_t) = \mathcal{F}_t
    \label{eq:disjoint-model-1}
\end{equation}
In the AVR literature, $f_*, f_\#, \ldots, f_\$$ are most often implemented as neural network components.
This leads to the following general form of a model for solving problems from $t$:
\begin{equation}
    \mathcal{F}_t (\{\bm{x}_{i,j}^t\}_{j=1}^{P_t}) = \bm{\widehat{y}}_i^t
    \label{eq:disjoint-model-2}
\end{equation}
Due to the dependence on both $\mathcal{S}_t$ (implicit) and $P_t$ (explicit), $\mathcal{F}_t$ cannot be directly applied to solving any task $t' \in \mathcal{T}^{'}_{t}$ with a different structure or number of panels: $\mathcal{T}^{'}_{t} = \{t' \in \mathcal{T} : \mathcal{S}_t \neq \mathcal{S}_{t'} \lor P_t \neq P_{t'} \}$

\subsection{Proposed new perspective (\textit{unified})}\label{sec:unified-perspective}
Instead of treating an AVR problem instance as a set of images with an associated structure, we propose to employ a rendering algorithm $\mathcal{R}$ that merges separate images, given the task's structure, into a single image $\chi \in [0, 1]^{h' \times w'}$ with new height $h'$ and width $w'$ (see Fig.~\ref{fig:intro}):
\begin{equation}
    \mathcal{R} (\{\bm{x}_{i,j}^t\}_{j=1}^{P_t} \mid \mathcal{S}_t) = \bm{\chi}_i^t
\end{equation}

In practice, $\mathcal{R}$ defines how instances of a given task should be presented.
To implement the algorithm for RPMs and VAPs considered in this work, for a given instance we first arrange the context panels, together with the missing panel (which is presented as a blank image with a centred question mark) into a grid with a small margin separating the panels.
This gives a $3\times3$ grid accommodating $8$ context panels for RPMs, and a $2\times3$ grid containing $5$ context panels for VAPs.
Next, we arrange the answer panels in another grid and position it below the context grid.
Depending on the task, the number of available answers ($n_a$) differs, e.g. matrices from VAP, G-set, and I-RAVEN datasets, have $n_a=4,5,8$, resp.
To accommodate this variability, we render the answer grid with up to $4$ panels in each row, resulting in up to $2$ rows.
A text label is placed above each answer panel.
Next, we initialize a blank canvas and resize the constructed image to fit the canvas.
While resizing, we keep the height to width ratio of the image in order to preserve the relative size of panel dimensions and to not distort the encompassed objects.
We fix the width of the canvas to $416$, and depending on the considered problem setting, set its height to $384$ for VAPs, $448$ for RPMs with $n_a \leq 4$, and $544$ for RPMs with $4 < n_a \leq 8$, which gives enough space to clearly render the panels.
All dimensions are divisible by $16$ to ensure that patch-based methods (with patch size $p = 16z$, for $z \in \mathbb{N}$) can be directly applied without the need to resize the underlying image.

$\mathcal{R}$ converts a considered AVR task into a common, unified representation that (a) facilitates construction of new universal AVR models capable of solving diverse AVR tasks, and (b) enables application of TL in existing SOTA AVR models. To the best of our knowledge, neither (a) nor (b) have ever been considered in the AVR literature. 
Within the above unified perspective, one can view any task $t$ as: $t = \{(\bm{\chi}_i^t, \bm{y}_i^t)\}_{i=1}^{N_t}$.

\subsection{General AVR solver}
Using the proposed unified perspective, it is possible to construct a general AVR model:
\vspace{-15pt}
\begin{center}
    \begin{tabular}{p{4.5cm}p{3cm}}
        \begin{equation}
            \mathcal{B}(f_* \circ f_\# \circ \ldots \circ f_\$) = \mathcal{F}
            \label{eq:unified-model-1}
        \end{equation}
        &
        \begin{equation}
            \mathcal{F} (\bm{\chi}_i^t) = \bm{\widehat{y}}_i
            \label{eq:unified-model-2}
        \end{equation}
        \\
    \end{tabular}
\end{center}
\vspace{-15pt}
Since $\mathcal{B}$ no longer depends on the task's structure (Eq.~\ref{eq:disjoint-model-1} vs Eq.~\ref{eq:unified-model-1}) and the model doesn't depend on the number of panels in the matrix (Eq.~\ref{eq:disjoint-model-2} vs Eq.~\ref{eq:unified-model-2}), the proposed unified perspective allows building a general model $\mathcal{F}$ applicable to solving diverse AVR tasks.
At the same time, $\mathcal{F}$ has to support input images $\chi_i^t$ that may vary in size.
Also, since $\mathcal{F}$ is no longer aware of which task it operates on, it has to provide its prediction in a common (fixed) format $\bm{\widehat{y}}$.
We assume that $\widehat{y} \in \mathbb{N}$ is an index of the correct answer, which is the case of RPMs, VAPs, and many other AVR tasks~\cite{malkinski2022review}.
This assumption may not be valid in some tasks, such as the original Bongard Problems~\cite{bongard1968recognition}, where an answer has to be provided in natural language.
An extension of the unified perspective to problems with specific output formats is left for future work.
Lastly, to be applicable to diverse AVR tasks, $\mathcal{F}$ has to be flexible enough to handle various structures of the tasks of interest.

\subsection{Proposed Unified AVR Model}\label{subsec:umavr}
In the initial experiments, we've discovered that SOTA CV baseline models struggle to deal with the above structural diversity.
Consequently, we propose UMAVR (Unified Model for Abstract Visual Reasoning) a neural architecture well-suited for the introduced unified view that takes rectangular images as input.
To build \emph{local} representations of low-level features, a convolutional backbone with depth $D_L$ is employed.
Each layer is composed of a convolution layer with kernel size $3\times3$, and $2\times2$ stride for dimensionality reduction, followed by Batch Normalization layer~\cite{ioffe2015batch} with ReLU activation~\cite{nair2010rectified}.
In the default setting, we use $D_L=4$, and the layers have $16$, $16$, $32$, and $128$ output channels, resp.
The size of the last channel determines the embedding size of a token at a given spatial location and is further referred to as $d$.
Applying this perception backbone yields a latent representation $z_0 \in \mathbb{R}^{d \times r \times c}$, where $r$ and $c$ are the numbers of rows and columns in the resultant token embedding matrix, resp.
For an image of size $544\times416$, this gives $z_0 \in \mathbb{R}^{128\times27\times25}$.

Next, we attach a component with a \emph{global} receptive field to facilitate discovery of patterns spanning multiple panels.
As a base layout, we adopt the MetaFormer architecture~\cite{yu2022metaformer}, which follows a layer-wise design, where the same layer is repeated $D_G$ times, and the weights are not shared across layers.
The operation performed in layer $l \in \left[1, D_G\right]$ can be formalized as:
\begin{align}
   z^*_l &= \text{TokenMixer}\left(\text{Norm}(z_{l-1})\right) + z_{l-1} \\
   z_l &= \text{ChannelMixer}\left(\text{Norm}(z^*_l)\right) + z^*_l
\end{align}
where $z_l$ is the output of the $l$'th layer, $z^*_l$ is an intermediate representation in layer $l$, and Norm is the Layer Normalization~\cite{ba2016layer}.
After the last layer, the token embedding matrix is passed through Layer Normalization, averaged along the width and height dimensions, and projected with a linear layer into $n_a-$dimensional vector.
The output is passed through the softmax function and the model is optimized end-to-end with cross-entropy.
The model architecture diagram is presented in Appendix~\ref{sec:umavr-architecture}.

\paragraph{TokenMixer.}
The input $z_{l-1}$ is normalized and passed through a residual 2D convolution layer with a kernel of size $5\times5$, and $2\times2$ padding to enrich tokens with the context from their spatial proximity.
Next, three parallel pathways are employed as introduced in the Vision Permutator (ViP)~\cite{hou2022vision}, which process the token matrix along the width, height, and channel dimensions, resp.
In each branch, $S=8$ segments are used.
Outputs of the pathways are concatenated depthwise and a 2D convolution with $1\times1$ kernel and $d$ output channels is applied to fuse information from the separate paths, in contrast to ViP which merges the branches through a summation or Split Attention~\cite{zhang2022resnest}.

\paragraph{ChannelMixer.}
The above global reasoning step is followed by local processing using a two-layer non-linear feed-forward block, input normalization, and a residual connection, as popularized by the Transformer~\cite{vaswani2017attention}: $ z_l = \sigma(\text{Norm}(z^*_l) W_1) W_2 + z^*_l,$
where $W_1 \in \mathbb{R}^{d \times kd}$ and $W_2 \in \mathbb{R}^{kd \times d}$ are learnable weights with an expansion factor $k$, and $\sigma$ is a non-linearity.
In the default setting, we use $k=4$ and the GELU activation~\cite{hendrycks2016gaussian}.

\section{Experiments}\label{sec:experiments}
We conduct experiments in three learning settings: STL, TL and CL. 
In each case, the performance of UMAVR is compared with the baseline models belonging to distinct model families.
All models are designed to return a logit vector $v \in \mathbb{R}^{n_a}$ representing a score for each answer, and the softmax function is used to compute the probability distribution $\widehat{p}$ over the set of answers.
The index corresponding to the highest probability is considered the predicted answer.

\paragraph{Baselines.}
The set of benchmark models includes convolutional networks, represented by ResNet~\cite{he2016deep} and ConvNext~\cite{liu2022convnet}, Vision Transformer (ViT)~\cite{dosovitskiy2021an}, MaxViT~\cite{tu2022maxvit}, TinyViT~\cite{wu2022tinyvit} and Swin Transformer~\cite{liu2021swin} as representatives of the Transformer family, adapted to vision tasks, and MLP-based models such as MLP-Mixer~\cite{tolstikhin2021mlp} and Vision Permutator (ViP)~\cite{hou2022vision}.

\begin{table}[t]
    \centering
    \caption{
    \textbf{Dataset details.}
    The benchmarks differ w.r.t. their size, allocation into train / val / test splits, and the maximal number of available answers $n_a^{max}$.
    }
    \label{tab:datasets}
    \begin{sc}
        \small
        \begin{tabular}{l|rrrr|c}
            \toprule
            Dataset & Size & Train & Val & Test & $n_a^{max}$ \\
            \midrule
            G-set & $49\text{K}$ & $34.4\text{K}$ & $9.8\text{K}$ & $4.8\text{K}$ & 5 \\
            I-RAVEN & $70\text{K}$ & $42\text{K}$ & $14\text{K}$ & $14\text{K}$ & 8 \\
            PGM & $1.42\text{M}$ & $1.2\text{M}$ & $20\text{K}$ & $200\text{K}$ & 8 \\
            VAP & $710\text{K}$ & $600\text{K}$ & $10\text{K}$ & $100\text{K}$ & 4 \\
            VASR & $154.8\text{K}$ & $150\text{K}$ & $2.25\text{K}$ & $2.55\text{K}$ & 4 \\
            \bottomrule
        \end{tabular}
    \end{sc}
\end{table}

\paragraph{Tasks.}
The models are evaluated on three challenging AVR problems.
Firstly, we consider the problem of solving RPMs from three datasets:
G-set~\cite{mandziuk2019deepiq,tomaszewska2022duel} with visually simple matrices following experiment~1 from~\cite{tomaszewska2022duel};
I-RAVEN~\cite{zhang2019raven,hu2021stratified} with a much richer set of available objects, attributes, and abstract rules;
and the Neutral regime of PGM~\cite{santoro2018measuring} which is of much bigger size.
Secondly, we consider the VAP dataset~\cite{hill2018learning} which presents a conceptual abstraction challenge with matrices that are structurally different from RPMs.
Thirdly, we employ the Visual Analogies of Situation Recognition dataset (VASR)~\cite{bitton2023vasr} that presents visual analogies comprising real-world images. We employ the dataset variant with random distractors.

Overall, the datasets vary in size, the content of the matrices (geometric shapes, real-world images), and the number of available answers (see Table~\ref{tab:datasets}).
This diversity allows gaining insights into models' performance in different axes.
To deeper analyse the limitations of the discussed models, in some experiments we reduced the number of available answers in the matrices, expecting that this simplification would make the task more comprehensible.

\paragraph{Experimental setting.}
The models are trained with batches of $256$ matrices for PGM and VAP datasets, and of $128$ matrices in all the remaining cases.
Model parameters are optimized with Adam~\cite{kingma2014adam} with $\beta_1=0.9$, $\beta_2=0.999$ and $\epsilon=10^{-8}$, until the validation loss doesn't improve for $10$ consecutive epochs.
We tuned learning rate $\lambda$ of each model separately with 3 randomly initialized runs on G-set and I-RAVEN with $n_a=2$ and selected $\lambda$ that worked best on average.
We applied linear learning rate warmup~\cite{liu2019variance} starting from $\lambda=10^{-6}$ over $500$ iterations and cosine decay~\cite{loshchilov2016sgdr} with $\lambda_{\text{min}}=10^{-6}$.
When learning to solve matrices from PGM, I-RAVEN and VAP, we make use of a supplementary training signal in the form of an auxiliary loss~\cite{santoro2018measuring,zhang2019raven} with sparse encoding~\cite{malkinski2020multi}, where the model has to additionally predict the hidden rules that govern the matrix construction.
Each training run is performed on a node with a single NVIDIA DGX A100 GPU.

In the experiments with $n_a < n_a^{max}$, in the original problem instance (with $n_a^{max}$ answer panels) the correct answer is preserved and then $n_a^{max}-n_a$ randomly sampled incorrect answers are deleted.

\renewcommand*{\numans}[1]{\parbox[c]{0.03\textwidth}{\centering #1}}%

\addtolength{\tabcolsep}{-1pt}

\begin{table*}[t]
    \centering
    \caption{
        \textbf{Single-task learning on abstract datasets.}
        Test accuracy of three baseline families of models (convolutional networks, vision transformers, MLP models for vision) and UMAVR in solving matrices from five datasets with variable numbers of possible answers $n_a$.
        Results higher than a random guess by more than $0.5\ \text{p.p}$ are highlighted with a blue background.
        Best results are marked in bold and the second best are underlined.
        P, N and T denote Pico, Nano and Tiny, resp.
    }
    \label{tab:single-task-learning}
    \begin{sc}
        \small
        \begin{tabular}{l|ccc|ccc|ccc|cc|c}
            \toprule
            \multirow{3}{*}{Model} & \multicolumn{12}{c}{Test accuracy (\%)} \\
            & \multicolumn{3}{c|}{G-set, $n_a=$} & \multicolumn{3}{c|}{I-RAVEN, $n_a=$} & \multicolumn{3}{c|}{PGM, $n_a=$} & \multicolumn{2}{c|}{VAP, $n_a=$} & VASR \\
            & \numans{$2$}          & \numans{$4$}          & \numans{$5$}          & \numans{$2$}          & \numans{$4$}          & \numans{$8$}       & \numans{$2$} & \numans{$4$} & \numans{$8$} & \numans{$2$} & \numans{$4$} & \numans{$4$} \\
            \midrule
            ResNet-18~\cite{he2016deep}               & \cl$97.0$             & \cl$91.5$             & \cl$91.1$             & $50.0$                & $25.0$                & $12.5$             & \cl$73.8$ & \cl$59.0$ & \cl$41.2$ & \cl$98.3$ & \cl$95.4$ & $25.0$ \\
            ResNet-50~\cite{he2016deep}               & \cl$96.9$             & \cl$94.1$             & \cl$81.7$                & $50.2$                & $25.0$                & $12.5$             & \cl$73.8$ & \cl$56.4$ & \cl$32.7$ & \cl$98.4$ & \cl$96.1$ & \cl$43.7$ \\
            ConvNext-P~\cite{liu2022convnet}  & $50.0$                & $25.0$                & $20.0$                & \cl$67.0$             & \cl$\underline{29.4}$ & $12.7$ & $50.0$ & $25.0$ & $12.5$ & \cl$96.0$ & \cl$90.4$ & $25.0$ \\
            ConvNext-N~\cite{liu2022convnet}  & $50.0$                & $25.0$                & $20.0$                & \cl$69.8$             & $25.1$                & $12.5$ & $50.0$ & $25.0$ & $12.5$ & \cl$95.9$ & \cl$91.7$ & $25.0$ \\
            \midrule
            MaxViT-P~\cite{tu2022maxvit}      & \cl$97.1$             & \cl$\underline{95.8}$ & \cl$\textbf{95.6}$ & $50.1$ & $25.2$ & $12.6$ & \cl$85.7$ & \cl$\textbf{97.1}$ & \cl$\underline{92.0}$ & \cl$\textbf{99.4}$ & \cl$\textbf{98.4}$ & \cl$\textbf{62.3}$ \\
            MaxViT-N~\cite{tu2022maxvit}      & \cl$\underline{97.3}$ & \cl$\textbf{96.1}$    & \cl$\underline{95.5}$ & \cl$81.0$ & $25.0$ & $12.7$ & \cl$85.2$ & \cl$\underline{94.6}$ & \cl$\textbf{97.3}$ & \cl$\underline{99.3}$ & \cl$\underline{98.2}$ & $24.0$ \\
            TinyViT-5M~\cite{wu2022tinyvit}      & \cl$97.2$             & \cl$95.6$             & \cl$95.4$ & \cl$83.5$ & $25.0$ & $12.5$ & \cl$72.7$ & \cl$41.1$ & \cl$34.3$ & \cl$99.0$ & \cl$97.1$ & \cl$55.4$ \\
            TinyViT-11M~\cite{wu2022tinyvit}     & \cl$97.0$             & \cl$94.7$             & \cl$95.4$ & \cl$71.2$ & $25.5$ & $12.7$ & \cl$84.3$ & \cl$46.6$ & \cl$34.0$ & \cl$98.3$ & \cl$96.3$ & \cl$54.2$ \\
            \midrule
            Mixer S/16~\cite{tolstikhin2021mlp}       & \cl$96.3$             & \cl$92.2$             & \cl$82.8$             & \cl$62.3$             & $25.5$ & $\underline{12.9}$ & \cl$\underline{88.8}$ & \cl$74.5$ & \cl$61.1$ & \cl$97.5$ & \cl$88.4$ & \cl$54.5$ \\
            Mixer S/32~\cite{tolstikhin2021mlp}       & \cl$97.0$             & \cl$95.6$             & \cl$94.9$             & \cl$74.9$             & $25.5$ & $12.8$ & \cl$\textbf{90.1}$ & \cl$76.4$ & \cl$73.4$ & \cl$97.2$ & \cl$93.0$ & \cl$54.1$ \\
            ViP-N~\cite{hou2022vision}        & $50.0$                & $25.0$                & $20.0$                & \cl$\underline{88.3}$ & $25.1$ & $12.5$ & \cl$77.6$ & \cl$46.3$ & \cl$81.7$ & \cl$97.6$ & \cl$95.2$ & $25.0$ \\
            ViP-T~\cite{hou2022vision}        & $50.0$                & $25.0$                & $20.0$                & \cl$75.6$             & $25.0$                & $12.5$             & \cl$85.2$ & \cl$73.8$ & \cl$49.0$ & \cl$97.3$ & \cl$95.5$ & $25.0$ \\
            \midrule
            UMAVR (ours) & \cl$\textbf{97.5}$    & \cl$\textbf{96.1}$    & \cl$95.1$             & \cl$\textbf{95.6}$    & \cl$\textbf{89.4}$ & \cl$\textbf{13.1}$ & \cl$76.9$ & \cl$63.3$ & \cl$52.3$ & \cl$93.9$ & \cl$97.3$ & \cl$\underline{59.8}$ \\
            \bottomrule
        \end{tabular}
    \end{sc}
\end{table*}

\addtolength{\tabcolsep}{1pt}

\subsection{Single-task learning}
\label{subsec:single-task-learning}
STL experiments assess the ability of modern CV models to solve uniformly viewed AVR tasks.
In preliminary experiments conducted on 4 AVR tasks (G-set, I-RAVEN, PGM, VAP) we discovered that large CV models typically struggled to perform better than chance, irrespectively of the dataset and the number of possible answers.
These models included ConvNext (Tiny, Small and Base), ViT-B/16, ViT-B/32, Swin (Tiny and Small), ViP-S/7 and ViP-M/7.
To overcome their limitations, in subsequent experiments we employed their smaller parameter-efficient variants including ConvNext Pico and Nano~\cite{rw2019timm}, MaxViT (Pico and Nano)~\cite{tu2022maxvit} and TinyViT (5M and 11M)~\cite{wu2022tinyvit}.
For ViP, variants smaller than Small weren't defined in the original paper, which lead us to construct two new variants coined Nano and Tiny.
Their detailed description is provided in Appendix~\ref{sec:experimental-details}.

Table~\ref{tab:single-task-learning} compares test accuracy of the models on 4 datasets with variable numbers of answers.
In G-set, where the amount of available data is scarce, ConvNext and ViP models struggle to perform better than chance, while remaining models achieve high performance, typically above $90\%$.
ResNet-50 and Mixer S/16 scored slightly above $80\%$ on the most challenging dataset configuration with $n_a=5$.
In I-RAVEN, which contains visually richer matrices with a hierarchical structure, both ResNet variants and MaxViT-Pico demonstrate performance at the random guess level across all $n_a$, other baseline models achieve non-random results only for $n_a=2$, while UMAVR significantly outcompetes all methods for $n_a \in \{2, 4\}$.
On PGM, all models but ConvNext learned to solve some matrices, though notably the best results were achieved by MaxViT ones.
On VAP, all models present satisfactory results, commonly exceeding $90\%$.
UMAVR demonstrates consistent and strong performance across nearly all considered settings.
The only exception is I-RAVEN with $n_a=8$, in which all tested models performed at the random guess level.
We conclude that UMAVR is a versatile method that in spite of its simplicity outcompetes other mainstream CV models in certain settings (specifically I-RAVEN with $n_a=4$).

\paragraph{Pre-trained checkpoints.}
To better understand the abstract reasoning capacity of large vision models, we repeated the STL experiments for ConvNext (Tiny, Small, Base), ViT (B/16, B/32) and Swin (Tiny, Small), starting from checkpoints pre-trained on ImageNet available in the TorchVision package~\cite{torchvision2016}.
However, the only setting where any of these models performed better than chance was I-RAVEN with $n_a=2$, where ConvNext-T and ConvNext-S achieved test accuracy of $75.6\%$ and $83.6\%$, resp.
This shows that despite using large pre-training datasets,
the contemporary large vision models are generally incapable of solving AVR tasks represented in a unified manner proposed in this paper.
Since the problem of classifying real-world images formulated in ImageNet is fundamentally different from AVR tasks considered in this work, we hypothesize that pre-training of large vision models on AVR data of sufficient scale might potentially boost their performance.
Until this hypothesis is validated in future work, relatively smaller, parameter-efficient supervised models remain SOTA in the domain.

\paragraph{Real-world analogies.}
The VASR dataset~\cite{bitton2023vasr} presents visual analogies formed from real-world images.
Methods used in~\cite{bitton2023vasr} employ pre-trained popular CV models (e.g. ViT or ConvNext) to embed each matrix image separately.
The answer is predicted by applying vector arithmetics~\cite{mikolov2013linguistic} or training a shallow supervised classifier on frozen image embeddings.
A limitation of these approaches is the lack of ability to reason about the relations between matrix panels in early layers of the model.
Instead, the reasoning is only framed as a post-processing step.
Differently, we applied the proposed unified view to VASR, which enables application of CV models in an entirely different setup, in which the models reason over the whole matrix starting already from early layers of the model.
The results are displayed in Table~\ref{tab:single-task-learning} (the rightmost column).
Certain models, including ResNet-18, both ConvNext variants, MaxVit-Nano and both ViP variants performed indistinguishably from random guessing.
Other models present performance typically exceeding $50\%$, with the best results achieved by MaxViT-Pico and UMAVR.
We conclude that VASR matrices presented in the unified view pose a significant challenge for the contemporary vision models.

\begin{table*}[t]
    \centering
    \caption{
        \textbf{Transfer learning.}
        Test accuracy of models that performed better than chance with STL on PGM (cf. Table~\ref{tab:single-task-learning}).
        The models are first pre-trained on PGM with $n_a=2,4,8$, resp., and then fine-tuned on G-set with $n_a=2,4,5$ and I-RAVEN with $n_a=2,4,8$, resp.
        In parentheses the difference w.r.t. STL training is presented.}
    \label{tab:transfer-learning}
    \begin{sc}
        \small
        \begin{tabular}{l|ccc|ccc}
            \toprule
            \multirow{2}{*}{Model} & \multicolumn{3}{c}{G-set test accuracy (\%)} & \multicolumn{3}{c}{I-RAVEN test accuracy (\%)} \\
            & $n_a=2$ & $n_a=4$ & $n_a=5$ & $n_a=2$ & $n_a=4$ & $n_a=8$ \\
            \midrule
            ResNet-18                             & \cl$96.7$ \scriptsize ($-\ \ 0.3$)                   & \cl$92.4$ \scriptsize ($+\ \ 0.9$) & \cl$93.0$ \scriptsize ($+\ \ 1.9$) & \cl$78.7$ \scriptsize ($+28.7$) & \cl$40.2$ \scriptsize ($+15.2$) & $12.5$ \scriptsize ($+\ \ 0.0$) \\
            ResNet-50                             & \cl$96.7$ \scriptsize ($-\ \ 0.2$)                            & \cl$92.9$ \scriptsize ($-\ \ 1.2$) & \cl$93.8$ \scriptsize ($+12.1$) & \cl$71.2$ \scriptsize ($+21.0$) & \cl$44.6$ \scriptsize ($+19.6$) & $12.5$ \scriptsize ($+\ \ 0.0$) \\
            \midrule
            MaxViT-P  & \cl$\underline{97.5}$ \scriptsize ($+\ \ 0.4$) & \cl$\underline{96.0}$ \scriptsize ($+\ \ 0.2$) & \cl$\underline{95.6}$ \scriptsize ($+\ \ 0.0$) & \cl$83.7$ \scriptsize ($+33.6$) & \cl$\underline{67.5}$ \scriptsize ($+42.3$) & \cl$41.6$ \scriptsize ($+29.0$) \\
            MaxViT-N  & \cl$97.2$ \scriptsize ($-\ \ 0.1$) & \cl$\textbf{96.2}$ \scriptsize ($+\ \ 0.1$) & \cl$\textbf{95.9}$ \scriptsize ($+\ \ 0.4$) & \cl$\underline{86.4}$ \scriptsize ($+\ \ 5.4$) & \cl$61.3$ \scriptsize ($+36.3$) & \cl$38.5$ \scriptsize ($+25.8$) \\
            TinyViT-5M  & \cl$95.7$ \scriptsize ($-\ \ 1.5$) & \cl$92.7$ \scriptsize ($-\ \ 2.9$) & \cl$92.9$ \scriptsize ($-\ \ 2.5$) & \cl$56.9$ \scriptsize ($-26.6$) & \cl$29.8$ \scriptsize ($+\ \ 4.8$) & \cl$26.0$ \scriptsize ($+13.5$) \\
            TinyViT-11M & \cl$97.4$ \scriptsize ($+\ \ 0.4$) & \cl$90.6$ \scriptsize ($-\ \ 4.1$) & \cl$\underline{95.6}$ \scriptsize ($+\ \ 0.2$) & \cl$82.7$ \scriptsize ($+11.5$) & \cl$46.6$ \scriptsize ($+21.1$) & \cl$27.3$ \scriptsize ($+14.6$) \\
            \midrule
            Mixer S/16   & \cl$96.9$ \scriptsize ($+\ \ 0.6$)                            & \cl$93.6$ \scriptsize ($+\ \ 1.4$) & \cl$91.6$ \scriptsize ($+\ \ 8.8$) & \cl$78.3$ \scriptsize ($+16.0$) & \cl$46.7$ \scriptsize ($+21.2$) & \cl$20.1$ \scriptsize ($+\ \ 7.2$) \\
            Mixer S/32   & \cl$96.5$ \scriptsize ($-\ \ 0.5$) & \cl$93.4$ \scriptsize ($-\ \ 2.2$) & \cl$94.9$ \scriptsize ($+\ \ 0.0$) & \cl$79.5$ \scriptsize ($+\ \ 4.6$) & \cl$45.4$ \scriptsize ($+19.9$) & \cl$36.2$ \scriptsize ($+23.4$) \\
            ViP-N    & \cl$96.3$ \scriptsize ($+46.3$)                               & \cl$91.8$ \scriptsize ($+66.8$) & \cl$95.0$ \scriptsize ($+75.0$) & \cl$74.0$ \scriptsize ($-14.3$) & \cl$46.4$ \scriptsize ($+21.3$) & \cl$\textbf{43.6}$ \scriptsize ($+31.1$) \\
            ViP-T    & \cl$\textbf{97.6}$ \scriptsize ($+47.6$)                               & \cl$95.2$ \scriptsize ($+70.2$) & \cl$94.9$ \scriptsize ($+74.9$) & \cl$81.2$ \scriptsize ($+\ \ 5.6$) & \cl$60.1$ \scriptsize ($+35.1$) & \cl$28.4$ \scriptsize ($+15.9$) \\
            \midrule
            UMAVR (ours) & \cl$97.3$ \scriptsize ($-\ \ 0.2$) & \cl$95.9$ \scriptsize ($-\ \ 0.2$) & \cl$95.4$ \scriptsize ($+\ \ 0.3$) & \cl$\textbf{95.5}$ \scriptsize ($-\ \ 0.1$) & \cl$\textbf{89.7}$ \scriptsize ($+\ \ 0.3$) & \cl$\underline{42.8}$ \scriptsize ($+29.7$) \\
            \bottomrule
        \end{tabular}
    \end{sc}
\end{table*}

\subsection{Transfer learning}
Next, for the models that performed better than chance in at least one of the STL experiments, we explore a TL scenario, where models pre-trained on the largest dataset (PGM) are fine-tuned on the two smallest ones (G-set or I-RAVEN).
For both the pre-training and fine-tuning datasets the same value of $n_a$ is used, except for G-set with $n_a=5$, for which pre-training on PGM is performed with $n_a=8$.
Table~\ref{tab:transfer-learning} presents the results.
Application of TL leads to significant gains in multiple considered settings.
Specifically, the performance of ResNet-50 and Mixer S/16 improved respectively from $81.7\%$ and $82.8\%$ to $93.8\%$ and $91.6\%$ on G-set with $n_a=5$, while the performance of ViP Nano and Tiny improved from random guessing level to being on-par with other top performers across all $n_a$ configurations.
On I-RAVEN, significant improvement across most considered settings is observed, as after TL only ResNets struggle in the most demanding setting ($n_a=8$).

The results signify the importance of utilizing a shared problem representation in the AVR domain, as pre-training the models on a large dataset can lead to notable performance improvements on the tasks, for which the available data is scarce.
In certain cases, however, we observe the impact of the negative transfer effect~\cite{pan2010survey,cao2018partial}, which brings attention to the need of designing robust TL techniques.

\subsection{Curriculum learning}
We evaluate the CL approach, in which the model is iteratively trained on gradually more demanding matrices (with bigger $n_a$), with reusing the previously gathered knowledge, see Algorithm~\ref{alg:curriculum-learning}.

\begin{algorithm}[t]
    \caption{\textbf{Curriculum learning.} Solving matrices of gradually increasing difficulty.}
    \label{alg:curriculum-learning}
    \begin{varwidth}[t]{\linewidth}
        \textbf{Input:} randomly initialized model $f_\theta$, $n_a^{\text{max}}$, $\lambda_0$\par
        \textbf{Output:} trained model $f_\theta$
    \end{varwidth}
    \begin{algorithmic}[1]
        \STATE $n_a = 2$
        \WHILE {$n_a \leq n_a^{max}$}
        \STATE $\lambda \gets \lambda_0$ \quad {\color{gray}\# reset the learning rate}\\
        {\color{gray}\# until convergence with early stopping}
        \STATE $f_\theta \gets$ optimize$(f_\theta, n_a, \lambda)$
        \STATE $n_a \gets n_a + 1$
        \ENDWHILE
    \end{algorithmic}
\end{algorithm}

We considered all models listed in Table~\ref{tab:single-task-learning} and evaluated them with CL on G-set and I-RAVEN.
On the former dataset, CL improved the performance of ResNet-50 from $81.7\%$ to $90.4\%$ and Mixer S/16 from $82.8\%$ to $95.2\%$.
On I-RAVEN, CL improved the results of MaxViT Pico to $75.7\%$ ($+63.1$ p.p.), TinyViT-5M to $28.3\%$ ($+15.8$ p.p.), TinyViT-11M to $31.4\%$ ($+18.7$ p.p.), Mixer S/16 to $31.7\%$ ($+18.8$ p.p.), Mixer S/32 to $37.7\%$ ($+24.9$ p.p.), ViP Tiny to $55.7\%$ ($+43.2$ p.p.), and UMAVR to $86.9\%$ ($+73.8$ p.p.).
Performance in the remaining settings stayed at the STL level ($\pm 0.3$ p.p.).
Overall, for certain models application of CL raised the results significantly compared to STL including the best-performing model -- UMAVR, showing the potential of effective knowledge reuse within the unified view framework.

\begin{figure*}[t]
    \centering
    \includegraphics[width=0.95\textwidth]{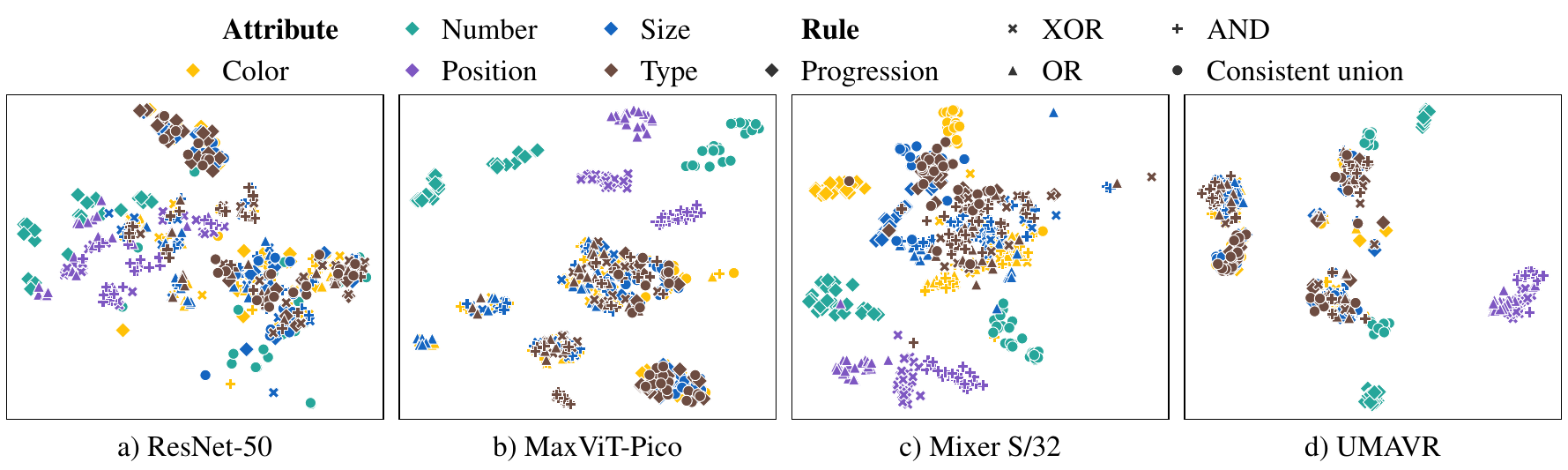}
    \caption{
    \textbf{PGM embeddings.}
    The embeddings of PGM matrices ($n_a=2$) from the test split of the \texttt{Neutral} regime, visualized with t-SNE~\cite{van2008visualizing}.
    For the sake of interpretability, the figure considers matrices with a single rule applied to Shape objects.
    }
    \label{fig:embeddings}
\end{figure*}

\subsection{Qualitative analysis}
Figure~\ref{fig:embeddings} compares UMAVR matrix embeddings ($n_a=2$) with selected representative models on PGM.
Overall, the clustering quality correlates with model's performance on the target task (cf. Table~\ref{tab:single-task-learning}).
Across all visualized models, the embeddings of matrices with rules applied to Number and Position attributes (green and purple, resp.) cluster into distinct groups.
In addition, the embeddings of Mixer S/32 start to form distinguishable clusters for the remaining attributes as well, which aligns with model's leading performance in this setting.
The visualization confirms that the models learn to identify the underlying abstract rules instead of relying on visual shortcuts or dataset biases.
Appendix~\ref{sec:embedding-visualization} extends this analysis to all relevant models and datasets (I-RAVEN, PGM and VAP).

\section{Conclusions and future work}\label{sec:conclusion}
Existing CV models that excel in solving AVR problems are generally task-specific, which prevents their application to other, even similar problems.
In this work, we have formulated a unified view on AVR tasks in which an AVR instance is regarded as a single image, with no indication about the location or role of individual panels (context panels vs. answer panels).
Apparently, even the SOTA CV models struggle to efficiently perform in this new setting, and in certain cases are unable to exceed a random guess level.
To address this new challenge, we propose the UMAVR model, applicable to solving diverse AVR tasks presented in the above unified perspective.
UMAVR shows very promising performance in the STL setup, and furthermore, demonstrates effective knowledge reuse in TL and CL setups, surpassing the performance of strong baselines.

The development of universal AVR methods has potential to foster progress in related areas via knowledge transfer. One possible target domain is document understanding~\cite{binmakhashen2019document} that requires a high degree of relational reasoning. Large-scale datasets, with uniformly viewed AVR tasks, could be used to pre-train effective reasoning models and facilitate the development of new solutions in this area.

\section*{Acknowledgements}
This research was carried out with the support of the Laboratory of Bioinformatics and Computational Genomics and the High Performance Computing Center of the Faculty of Mathematics and Information Science Warsaw University of Technology.
Mikołaj Małkiński was funded by the Warsaw University of Technology within the Excellence Initiative: Research University (IDUB) programme.

{
\small
\bibliography{main}

\begin{thebibliography}{55}
\providecommand{\natexlab}[1]{#1}

\bibitem[{Ba, Kiros, and Hinton(2016)}]{ba2016layer}
Ba, J.~L.; Kiros, J.~R.; and Hinton, G.~E. 2016.
\newblock Layer normalization.
\newblock \emph{arXiv preprint arXiv:1607.06450}.

\bibitem[{Barrett et~al.(2018)Barrett, Hill, Santoro, Morcos, and Lillicrap}]{santoro2018measuring}
Barrett, D.; Hill, F.; Santoro, A.; Morcos, A.; and Lillicrap, T. 2018.
\newblock Measuring abstract reasoning in neural networks.
\newblock In \emph{International Conference on Machine Learning}, 511--520. PMLR.

\bibitem[{Bengio et~al.(2009)Bengio, Louradour, Collobert, and Weston}]{bengio2009curriculum}
Bengio, Y.; Louradour, J.; Collobert, R.; and Weston, J. 2009.
\newblock Curriculum learning.
\newblock In \emph{Proceedings of the 26th annual international conference on machine learning}, 41--48.

\bibitem[{Binmakhashen and Mahmoud(2019)}]{binmakhashen2019document}
Binmakhashen, G.~M.; and Mahmoud, S.~A. 2019.
\newblock Document layout analysis: a comprehensive survey.
\newblock \emph{ACM Computing Surveys (CSUR)}, 52(6): 1--36.

\bibitem[{Bitton et~al.(2023)Bitton, Yosef, Strugo, Shahaf, Schwartz, and Stanovsky}]{bitton2023vasr}
Bitton, Y.; Yosef, R.; Strugo, E.; Shahaf, D.; Schwartz, R.; and Stanovsky, G. 2023.
\newblock {VASR}: Visual analogies of situation recognition.
\newblock In \emph{Proceedings of the AAAI Conference on Artificial Intelligence}, volume~37, 241--249.

\bibitem[{Bongard(1968)}]{bongard1968recognition}
Bongard, M.~M. 1968.
\newblock The recognition problem.
\newblock Technical report, Foreign Technology Div Wright-Patterson AFB Ohio.

\bibitem[{Cao et~al.(2018)Cao, Long, Wang, and Jordan}]{cao2018partial}
Cao, Z.; Long, M.; Wang, J.; and Jordan, M.~I. 2018.
\newblock Partial transfer learning with selective adversarial networks.
\newblock In \emph{Proceedings of the IEEE conference on computer vision and pattern recognition}, 2724--2732.

\bibitem[{Carpenter, Just, and Shell(1990)}]{carpenter1990one}
Carpenter, P.~A.; Just, M.~A.; and Shell, P. 1990.
\newblock What one intelligence test measures: a theoretical account of the processing in the Raven Progressive Matrices Test.
\newblock \emph{Psychological review}, 97(3): 404.

\bibitem[{Dosovitskiy et~al.(2021)Dosovitskiy, Beyer, Kolesnikov, Weissenborn, Zhai, Unterthiner, Dehghani, Minderer, Heigold, Gelly, Uszkoreit, and Houlsby}]{dosovitskiy2021an}
Dosovitskiy, A.; Beyer, L.; Kolesnikov, A.; Weissenborn, D.; Zhai, X.; Unterthiner, T.; Dehghani, M.; Minderer, M.; Heigold, G.; Gelly, S.; Uszkoreit, J.; and Houlsby, N. 2021.
\newblock An Image is Worth 16x16 Words: Transformers for Image Recognition at Scale.
\newblock In \emph{International Conference on Learning Representations}.

\bibitem[{He et~al.(2016)He, Zhang, Ren, and Sun}]{he2016deep}
He, K.; Zhang, X.; Ren, S.; and Sun, J. 2016.
\newblock Deep residual learning for image recognition.
\newblock In \emph{Proceedings of the IEEE conference on computer vision and pattern recognition}, 770--778.

\bibitem[{Hendrycks and Gimpel(2016)}]{hendrycks2016gaussian}
Hendrycks, D.; and Gimpel, K. 2016.
\newblock Gaussian error linear units (gelus).
\newblock \emph{arXiv preprint arXiv:1606.08415}.

\bibitem[{Hill et~al.(2019)Hill, Santoro, Barrett, Morcos, and Lillicrap}]{hill2018learning}
Hill, F.; Santoro, A.; Barrett, D.; Morcos, A.; and Lillicrap, T. 2019.
\newblock Learning to Make Analogies by Contrasting Abstract Relational Structure.
\newblock In \emph{International Conference on Learning Representations}.

\bibitem[{Hinton, Krizhevsky, and Sutskever(2012)}]{hinton2012imagenet}
Hinton, G.~E.; Krizhevsky, A.; and Sutskever, I. 2012.
\newblock Imagenet classification with deep convolutional neural networks.
\newblock \emph{Advances in neural information processing systems}, 25(1106-1114): 1.

\bibitem[{Hoshen and Werman(2017)}]{hoshen2017iq}
Hoshen, D.; and Werman, M. 2017.
\newblock {IQ} of neural networks.
\newblock \emph{arXiv preprint arXiv:1710.01692}.

\bibitem[{Hou et~al.(2022)Hou, Jiang, Yuan, Cheng, Yan, and Feng}]{hou2022vision}
Hou, Q.; Jiang, Z.; Yuan, L.; Cheng, M.-M.; Yan, S.; and Feng, J. 2022.
\newblock Vision permutator: A permutable mlp-like architecture for visual recognition.
\newblock \emph{IEEE Transactions on Pattern Analysis and Machine Intelligence}.

\bibitem[{Hu et~al.(2021)Hu, Ma, Liu, Wei, and Bai}]{hu2021stratified}
Hu, S.; Ma, Y.; Liu, X.; Wei, Y.; and Bai, S. 2021.
\newblock Stratified Rule-Aware Network for Abstract Visual Reasoning.
\newblock In \emph{Proceedings of the AAAI Conference on Artificial Intelligence}, volume~35, 1567--1574.

\bibitem[{Ioffe and Szegedy(2015)}]{ioffe2015batch}
Ioffe, S.; and Szegedy, C. 2015.
\newblock Batch normalization: Accelerating deep network training by reducing internal covariate shift.
\newblock In \emph{International Conference on Machine Learning}, 448--456. PMLR.

\bibitem[{Kingma and Ba(2014)}]{kingma2014adam}
Kingma, D.~P.; and Ba, J. 2014.
\newblock Adam: A method for stochastic optimization.
\newblock \emph{International Conference on Learning Representations}.

\bibitem[{Liu et~al.(2020)Liu, Jiang, He, Chen, Liu, Gao, and Han}]{liu2019variance}
Liu, L.; Jiang, H.; He, P.; Chen, W.; Liu, X.; Gao, J.; and Han, J. 2020.
\newblock On the Variance of the Adaptive Learning Rate and Beyond.
\newblock In \emph{International Conference on Learning Representations}.

\bibitem[{Liu et~al.(2021)Liu, Lin, Cao, Hu, Wei, Zhang, Lin, and Guo}]{liu2021swin}
Liu, Z.; Lin, Y.; Cao, Y.; Hu, H.; Wei, Y.; Zhang, Z.; Lin, S.; and Guo, B. 2021.
\newblock Swin transformer: Hierarchical vision transformer using shifted windows.
\newblock In \emph{Proceedings of the IEEE/CVF International Conference on Computer Vision}, 10012--10022.

\bibitem[{Liu et~al.(2022)Liu, Mao, Wu, Feichtenhofer, Darrell, and Xie}]{liu2022convnet}
Liu, Z.; Mao, H.; Wu, C.-Y.; Feichtenhofer, C.; Darrell, T.; and Xie, S. 2022.
\newblock A convnet for the 2020s.
\newblock In \emph{Proceedings of the IEEE/CVF Conference on Computer Vision and Pattern Recognition}, 11976--11986.

\bibitem[{Locatello et~al.(2020)Locatello, Weissenborn, Unterthiner, Mahendran, Heigold, Uszkoreit, Dosovitskiy, and Kipf}]{locatello2020object}
Locatello, F.; Weissenborn, D.; Unterthiner, T.; Mahendran, A.; Heigold, G.; Uszkoreit, J.; Dosovitskiy, A.; and Kipf, T. 2020.
\newblock Object-centric learning with slot attention.
\newblock \emph{Advances in Neural Information Processing Systems}, 33: 11525--11538.

\bibitem[{Loshchilov and Hutter(2017)}]{loshchilov2016sgdr}
Loshchilov, I.; and Hutter, F. 2017.
\newblock {SGDR}: Stochastic Gradient Descent with Warm Restarts.
\newblock In \emph{International Conference on Learning Representations}.

\bibitem[{maintainers and contributors(2016)}]{torchvision2016}
maintainers, T.; and contributors. 2016.
\newblock TorchVision: PyTorch's Computer Vision library.
\newblock \url{https://github.com/pytorch/vision}.

\bibitem[{Ma{\l}ki{\'n}ski and Ma{\'n}dziuk(2022)}]{malkinski2022deep}
Ma{\l}ki{\'n}ski, M.; and Ma{\'n}dziuk, J. 2022.
\newblock Deep Learning Methods for Abstract Visual Reasoning: A Survey on Raven's Progressive Matrices.
\newblock \emph{arXiv preprint arXiv:2201.12382}.

\bibitem[{Ma{\l}ki{\'n}ski and Ma{\'n}dziuk(2023)}]{malkinski2022review}
Ma{\l}ki{\'n}ski, M.; and Ma{\'n}dziuk, J. 2023.
\newblock A Review of Emerging Research Directions in Abstract Visual Reasoning.
\newblock \emph{Information Fusion}, 91: 713--736.

\bibitem[{Ma{\l}ki{\'n}ski and Ma{\'n}dziuk(2024{\natexlab{a}})}]{malkinski2020multi}
Ma{\l}ki{\'n}ski, M.; and Ma{\'n}dziuk, J. 2024{\natexlab{a}}.
\newblock Multi-Label Contrastive Learning for Abstract Visual Reasoning.
\newblock \emph{IEEE Transactions on Neural Networks and Learning Systems}, 35(2): 1941--1953.

\bibitem[{Ma{\l}ki{\'n}ski and Ma{\'n}dziuk(2024{\natexlab{b}})}]{malkinski2024one}
Ma{\l}ki{\'n}ski, M.; and Ma{\'n}dziuk, J. 2024{\natexlab{b}}.
\newblock One Self-Configurable Model to Solve Many Abstract Visual Reasoning Problems.
\newblock In \emph{Proceedings of the AAAI Conference on Artificial Intelligence}, volume~38, 14297--14305.

\bibitem[{Ma{\'n}dziuk and {\.Z}ychowski(2019)}]{mandziuk2019deepiq}
Ma{\'n}dziuk, J.; and {\.Z}ychowski, A. 2019.
\newblock {DeepIQ}: A Human-Inspired {AI} System for Solving {IQ} Test Problems.
\newblock In \emph{2019 International Joint Conference on Neural Networks}, 1--8. IEEE.

\bibitem[{Mikolov, Yih, and Zweig(2013)}]{mikolov2013linguistic}
Mikolov, T.; Yih, W.-t.; and Zweig, G. 2013.
\newblock Linguistic regularities in continuous space word representations.
\newblock In \emph{Proceedings of the 2013 conference of the north american chapter of the association for computational linguistics: Human language technologies}, 746--751.

\bibitem[{Mondal, Webb, and Cohen(2023)}]{mondal2023learning}
Mondal, S.~S.; Webb, T.~W.; and Cohen, J. 2023.
\newblock Learning to reason over visual objects.
\newblock In \emph{The Eleventh International Conference on Learning Representations}.

\bibitem[{Nair and Hinton(2010)}]{nair2010rectified}
Nair, V.; and Hinton, G.~E. 2010.
\newblock Rectified linear units improve restricted boltzmann machines.
\newblock In \emph{Proceedings of the 27th International Conference on International Conference on Machine Learning}, 807–814. Omnipress.

\bibitem[{Nie et~al.(2020)Nie, Yu, Mao, Patel, Zhu, and Anandkumar}]{nie2020bongard}
Nie, W.; Yu, Z.; Mao, L.; Patel, A.~B.; Zhu, Y.; and Anandkumar, A. 2020.
\newblock Bongard-logo: A new benchmark for human-level concept learning and reasoning.
\newblock \emph{Advances in Neural Information Processing Systems}, 33: 16468--16480.

\bibitem[{Pan and Yang(2010)}]{pan2010survey}
Pan, S.~J.; and Yang, Q. 2010.
\newblock A survey on transfer learning.
\newblock \emph{IEEE Transactions on knowledge and data engineering}, 22(10): 1345--1359.

\bibitem[{Raven(1936)}]{raven1936mental}
Raven, J.~C. 1936.
\newblock Mental tests used in genetic studies: The performance of related individuals on tests mainly educative and mainly reproductive.
\newblock \emph{Master’s thesis, University of London}.

\bibitem[{Raven and Court(1998)}]{raven1998raven}
Raven, J.~C.; and Court, J.~H. 1998.
\newblock \emph{Raven's progressive matrices and vocabulary scales}.
\newblock Oxford pyschologists Press Oxford, England.

\bibitem[{Rogozhnikov(2022)}]{rogozhnikov2022einops}
Rogozhnikov, A. 2022.
\newblock Einops: Clear and Reliable Tensor Manipulations with Einstein-like Notation.
\newblock In \emph{International Conference on Learning Representations}.

\bibitem[{Santoro et~al.(2017)Santoro, Raposo, Barrett, Malinowski, Pascanu, Battaglia, and Lillicrap}]{santoro2017simple}
Santoro, A.; Raposo, D.; Barrett, D.~G.; Malinowski, M.; Pascanu, R.; Battaglia, P.; and Lillicrap, T. 2017.
\newblock A simple neural network module for relational reasoning.
\newblock \emph{Advances in neural information processing systems}, 30: 4967--4976.

\bibitem[{Snow, Kyllonen, and Marshalek(1984)}]{snow1984topography}
Snow, R.~E.; Kyllonen, P.~C.; and Marshalek, B. 1984.
\newblock {The topography of ability and learning correlations}.
\newblock \emph{Advances in the psychology of human intelligence}, 2(S 47): 103.

\bibitem[{Tolstikhin et~al.(2021)Tolstikhin, Houlsby, Kolesnikov, Beyer, Zhai, Unterthiner, Yung, Steiner, Keysers, Uszkoreit et~al.}]{tolstikhin2021mlp}
Tolstikhin, I.~O.; Houlsby, N.; Kolesnikov, A.; Beyer, L.; Zhai, X.; Unterthiner, T.; Yung, J.; Steiner, A.; Keysers, D.; Uszkoreit, J.; et~al. 2021.
\newblock Mlp-mixer: An all-mlp architecture for vision.
\newblock \emph{Advances in Neural Information Processing Systems}, 34: 24261--24272.

\bibitem[{Tomaszewska, {\.Z}ychowski, and Ma{\'n}dziuk(2022)}]{tomaszewska2022duel}
Tomaszewska, P.; {\.Z}ychowski, A.; and Ma{\'n}dziuk, J. 2022.
\newblock Duel-based Deep Learning system for solving IQ tests.
\newblock In \emph{International Conference on Artificial Intelligence and Statistics}, 10483--10492. PMLR.

\bibitem[{Triantafillou et~al.(2020)Triantafillou, Zhu, Dumoulin, Lamblin, Evci, Xu, Goroshin, Gelada, Swersky, Manzagol, and Larochelle}]{triantafillou2020meta}
Triantafillou, E.; Zhu, T.; Dumoulin, V.; Lamblin, P.; Evci, U.; Xu, K.; Goroshin, R.; Gelada, C.; Swersky, K.; Manzagol, P.-A.; and Larochelle, H. 2020.
\newblock Meta-Dataset: A Dataset of Datasets for Learning to Learn from Few Examples.
\newblock In \emph{International Conference on Learning Representations}.

\bibitem[{Tu et~al.(2022)Tu, Talebi, Zhang, Yang, Milanfar, Bovik, and Li}]{tu2022maxvit}
Tu, Z.; Talebi, H.; Zhang, H.; Yang, F.; Milanfar, P.; Bovik, A.; and Li, Y. 2022.
\newblock Maxvit: Multi-axis vision transformer.
\newblock In \emph{European conference on computer vision}, 459--479. Springer.

\bibitem[{Van~der Maaten and Hinton(2008)}]{van2008visualizing}
Van~der Maaten, L.; and Hinton, G. 2008.
\newblock Visualizing data using t-SNE.
\newblock \emph{Journal of machine learning research}, 9(11).

\bibitem[{Vaswani et~al.(2017)Vaswani, Shazeer, Parmar, Uszkoreit, Jones, Gomez, Kaiser, and Polosukhin}]{vaswani2017attention}
Vaswani, A.; Shazeer, N.; Parmar, N.; Uszkoreit, J.; Jones, L.; Gomez, A.~N.; Kaiser, {\L}.; and Polosukhin, I. 2017.
\newblock Attention is all you need.
\newblock \emph{Advances in neural information processing systems}, 30: 5998--6008.

\bibitem[{Wang, Jamnik, and Lio(2020)}]{wang2020abstract}
Wang, D.; Jamnik, M.; and Lio, P. 2020.
\newblock Abstract diagrammatic reasoning with multiplex graph networks.
\newblock In \emph{International Conference on Learning Representations}.

\bibitem[{Webb et~al.(2020)Webb, Dulberg, Frankland, Petrov, O’Reilly, and Cohen}]{webb2020learning}
Webb, T.; Dulberg, Z.; Frankland, S.; Petrov, A.; O’Reilly, R.; and Cohen, J. 2020.
\newblock Learning representations that support extrapolation.
\newblock In \emph{International Conference on Machine Learning}, 10136--10146. PMLR.

\bibitem[{Wightman(2019)}]{rw2019timm}
Wightman, R. 2019.
\newblock PyTorch Image Models.
\newblock \url{https://github.com/rwightman/pytorch-image-models}.

\bibitem[{Wu et~al.(2022)Wu, Zhang, Peng, Liu, Xiao, Fu, and Yuan}]{wu2022tinyvit}
Wu, K.; Zhang, J.; Peng, H.; Liu, M.; Xiao, B.; Fu, J.; and Yuan, L. 2022.
\newblock Tinyvit: Fast pretraining distillation for small vision transformers.
\newblock In \emph{European Conference on Computer Vision}, 68--85. Springer.

\bibitem[{Wu et~al.(2020)Wu, Dong, Grosse, and Ba}]{wu2020scattering}
Wu, Y.; Dong, H.; Grosse, R.; and Ba, J. 2020.
\newblock The Scattering Compositional Learner: Discovering Objects, Attributes, Relationships in Analogical Reasoning.
\newblock \emph{arXiv preprint arXiv:2007.04212}.

\bibitem[{Yu et~al.(2022)Yu, Luo, Zhou, Si, Zhou, Wang, Feng, and Yan}]{yu2022metaformer}
Yu, W.; Luo, M.; Zhou, P.; Si, C.; Zhou, Y.; Wang, X.; Feng, J.; and Yan, S. 2022.
\newblock Metaformer is actually what you need for vision.
\newblock In \emph{Proceedings of the IEEE/CVF Conference on Computer Vision and Pattern Recognition}, 10819--10829.

\bibitem[{Zhai et~al.(2019)Zhai, Puigcerver, Kolesnikov, Ruyssen, Riquelme, Lucic, Djolonga, Pinto, Neumann, Dosovitskiy et~al.}]{zhai2019large}
Zhai, X.; Puigcerver, J.; Kolesnikov, A.; Ruyssen, P.; Riquelme, C.; Lucic, M.; Djolonga, J.; Pinto, A.~S.; Neumann, M.; Dosovitskiy, A.; et~al. 2019.
\newblock A large-scale study of representation learning with the visual task adaptation benchmark.
\newblock \emph{arXiv preprint arXiv:1910.04867}.

\bibitem[{Zhang et~al.(2019)Zhang, Gao, Jia, Zhu, and Zhu}]{zhang2019raven}
Zhang, C.; Gao, F.; Jia, B.; Zhu, Y.; and Zhu, S.-C. 2019.
\newblock {RAVEN}: A dataset for relational and analogical visual reasoning.
\newblock In \emph{Proceedings of the IEEE/CVF Conference on Computer Vision and Pattern Recognition}, 5317--5327.

\bibitem[{Zhang et~al.(2022)Zhang, Wu, Zhang, Zhu, Lin, Zhang, Sun, He, Mueller, Manmatha et~al.}]{zhang2022resnest}
Zhang, H.; Wu, C.; Zhang, Z.; Zhu, Y.; Lin, H.; Zhang, Z.; Sun, Y.; He, T.; Mueller, J.; Manmatha, R.; et~al. 2022.
\newblock Resnest: Split-attention networks.
\newblock In \emph{Proceedings of the IEEE/CVF Conference on Computer Vision and Pattern Recognition}, 2736--2746.

\bibitem[{Zhang et~al.(2020)Zhang, Zhang, Zhu, and Zhu}]{zhang2020machine}
Zhang, W.; Zhang, C.; Zhu, Y.; and Zhu, S.-C. 2020.
\newblock Machine number sense: A dataset of visual arithmetic problems for abstract and relational reasoning.
\newblock In \emph{Proceedings of the AAAI Conference on Artificial Intelligence}, volume~34, 1332--1340.

\end{thebibliography}
}

\newpage
\appendix
\onecolumn

\section{Example matrices}\label{sec:examples}
Samples of matrices from the datasets considered in the paper are demonstrated in Figs.~\ref{fig:rpm-iraven}--\ref{fig:vasr}.

\section{Experimental details}\label{sec:experimental-details}

\paragraph{Auxiliary training.}
As mentioned in the paper, when learning to solve matrices from PGM, I-RAVEN and VAP, a supplementary training signal in the form of an auxiliary loss is utilized (Section~\ref{sec:experiments}).
To this end, in parallel to the answer prediction layer (Section~\ref{subsec:umavr}), a shallow rule classifier is applied.
The classifier operates on the token embedding matrix $z_{D_G}$ passed through the Layer Normalization and averaged along the width and height dimensions.
The classifier is composed of a linear layer with $128$ units, followed by GELU and another linear layer with $\vert r \vert$ output neurons, where $\vert r \vert$ is the size of the one-hot encoded rule vector.
Depending on the dataset, this gives $50$, $40$ and $28$ units for PGM, I-RAVEN and VAP, respectively, which corresponds to the number of unique rules in each dataset.

\paragraph{ViP variants.}
We propose 2 new parameter-efficient variants of ViP: Nano and Tiny.
Their specifications are listed in Table~\ref{tab:vip}.

\paragraph{Model size.}
Table~\ref{tab:model-size} compares the numbers of parameters of the models from Table~\ref{tab:single-task-learning}.

\begin{table}[H]
    \centering
    \caption{
    \textbf{Custom ViP variants.}
     We present two new variants of Vision Permutator (Nano and Tiny) following the same format as in~\cite{hou2022vision}.
     Since the number of tokens depends on the size of the input image, below we provide values for an image of size $448 \times 416$.}
    \label{tab:vip}
    \small
    \begin{tabular}{l|cc}
        \toprule
        Specification & ViP-Nano & ViP-Tiny \\
        \midrule
        Patch size & $16 \times 16$ & $16 \times 16$ \\
        Hidden size & - & $128$ \\
        \#Tokens & $28 \times 26$ & $28 \times 26$ \\
        \#Permutators & - & $2$ \\
        \midrule
        Patch size & - & $2 \times 2$ \\
        Hidden size & $128$ & $256$ \\
        \#Tokens & $28 \times 26$ & $14 \times 13$ \\
        \#Permutators & $8$ & $8$ \\
        \midrule
        Number of layers & $8$ & $10$ \\
        MLP Ratio & $2$ & $2$ \\
        Stoch. Dep. & $0.0$ & $0.0$ \\
        Parameters (M) & $4\text{M}$ & $7.9\text{M}$ \\
        \bottomrule
    \end{tabular}
\end{table}

\begin{table}[H]
    \centering
    \caption{\textbf{Model size.}
    The number of parameters of each model in millions (M).}
    \label{tab:model-size}
    \begin{sc}
        \small
        \begin{tabular}{l|r}
            \toprule
            Model & \# Parameters \\
            \midrule
            ResNet-18 & $11.2$M \\
            ResNet-50 & $23.8$M \\
            ConvNext-P & $8.7$M \\
            ConvNext-N & $15.1$M \\
            \midrule
            MaxViT-P & $7.3$M \\
            MaxViT-N & $15.0$M \\
            TinyViT-5M & $5.1$M \\
            TinyViT-11M & $10.6$M \\
            \midrule
            Mixer S/16 & $20.0$M \\
            Mixer S/32 & $18.2$M \\
            ViP-N & $4.0$M \\
            ViP-T & $7.9$M \\
            \midrule
            UMAVR & $3.5$M \\
            \bottomrule
        \end{tabular}
    \end{sc}
\end{table}

\definecolor{layer}{HTML}{E1E1E1}
\definecolor{activation}{HTML}{E69F00}
\definecolor{conv}{HTML}{56B4E9}
\definecolor{linear}{HTML}{009E73}
\definecolor{layernorm}{HTML}{CC79A7}
\definecolor{batchnorm}{HTML}{CC79A7}
\definecolor{channelmixer}{HTML}{F0E442}
\definecolor{tokenmixer}{HTML}{D55E00}

\tikzstyle{layer} = [rectangle, rounded corners=2pt, text centered, draw=black, fill=layer, inner sep=0, minimum height=1.2*\unit, minimum width=6*\unit]
\tikzstyle{label} = [align=center, font={\footnotesize}]
\tikzstyle{panellabel} = [rectangle, inner sep=0]
\tikzstyle{rearrange} = [layer, inner sep=2pt, align=center]
\tikzstyle{activation} = [layer, fill=activation!50]
\tikzstyle{conv2d} = [layer, fill=conv!50]
\tikzstyle{linear} = [layer, fill=linear!50]
\tikzstyle{layernorm} = [layer, fill=layernorm!20]
\tikzstyle{batchnorm} = [layer, fill=batchnorm!20]
\tikzstyle{channelmixer} = [layer, fill=channelmixer!10]
\tikzstyle{tokenmixer} = [layer, fill=tokenmixer!10]
\tikzstyle{resplus} = [circle, draw=black, inner sep=0]
\tikzstyle{emptylayer} = [rectangle, inner sep=0, minimum height=1.2*\unit, minimum width=6*\unit]
\tikzstyle{background} = [draw, fill=black!5, rounded corners=5pt, densely dashed, inner sep=0]

\tikzstyle{embedding} = [rectangle, rounded corners=7pt, text centered, draw=black, fill=layer!50, inner sep=0, minimum height=1.6*\unit, minimum width=2*\unit]
\tikzstyle{panel} = [thick, rectangle, draw=black, inner sep=0]

\tikzstyle{arrow} = [-Stealth]

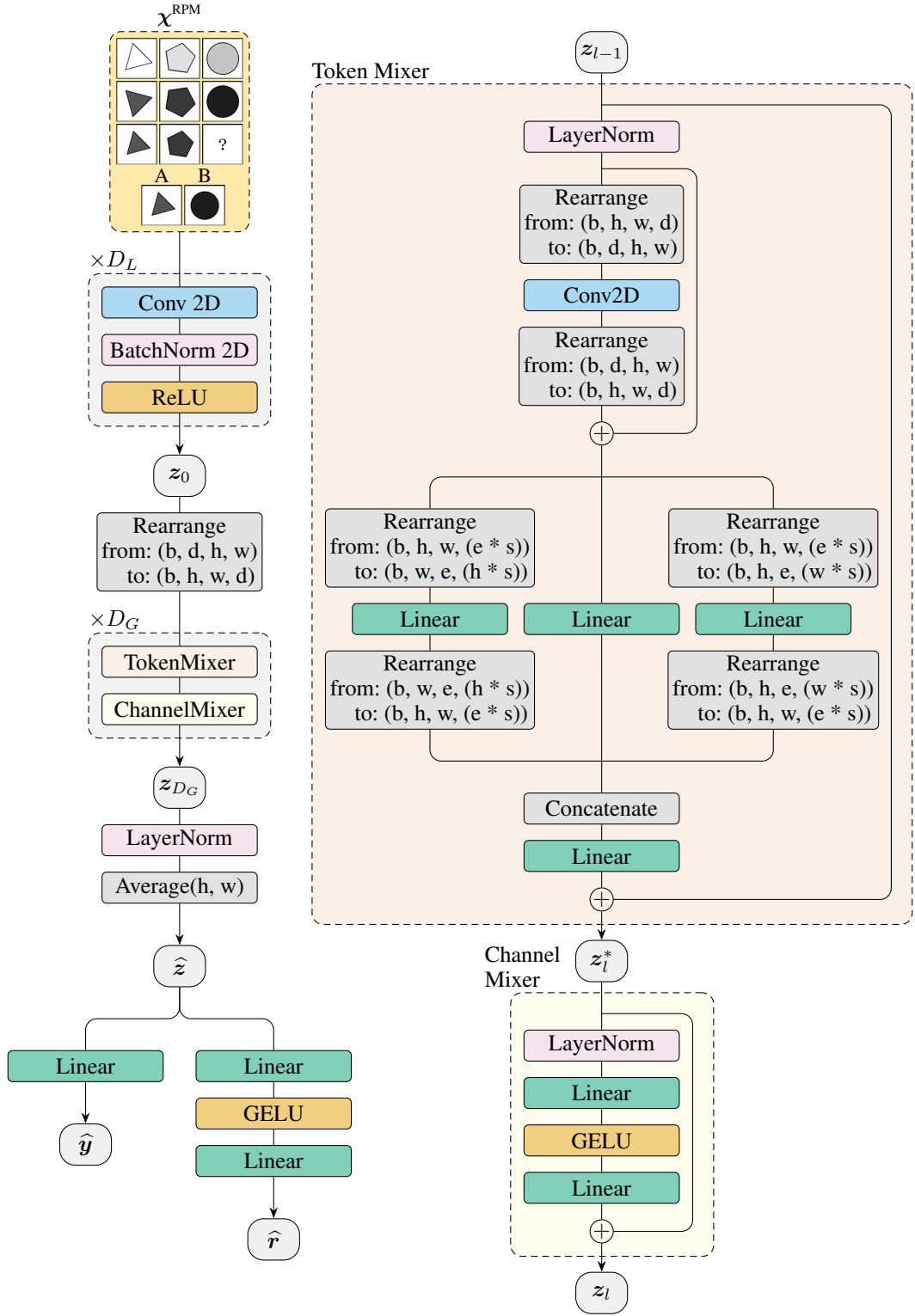
\begin{figure}[t]
    \centering
    \begin{tikzpicture}


        \begin{scope}[node distance=1pt]
            \node (r1) [panel] {\rpm{0}};
            \node (r2) [panel, right=of r1] {\rpm{1}};
            \node (r3) [panel, right=of r2] {\rpm{2}};
            \node (r4) [panel, below=of r1] {\rpm{3}};
            \node (r5) [panel, below=of r2] {\rpm{4}};
            \node (r6) [panel, below=of r3] {\rpm{5}};
            \node (r7) [panel, below=of r4] {\rpm{6}};
            \node (r8) [panel, below=of r5] {\rpm{7}};
            \node (r9) [panel, below=of r6] {\rpm{8}};
            \node (ra1) [panel, below=of r7, xshift=11pt, yshift=-8pt] {\rpmans{2}};
            \node (ra1t) [panellabel, above=of ra1] {\footnotesize A};
            \node (ra2) [panel, right=of ra1] {\rpmans{7}};
            \node (ra2t) [panellabel, above=of ra2] {\footnotesize B};
        \end{scope}
        \begin{scope}[on background layer]
            \node (rbg) [fit={(r1.north west) (r3.east) (ra1.south)}, background, fill=rpm!35, inner sep=0.25*\unit] {};
        \end{scope}
        \node (rut) [inner sep=0pt] at ($ (rbg.north) + (0, 0.6*\unit) $) {\small $\bm{\chi}^{\text{RPM}}$};

        \begin{scope}[node distance=0.60*\unit]
            \node (con1) [conv2d, below=of rbg, yshift=-1.5*\unit] {Conv 2D};
            \node (bn1) [batchnorm, below=of con1] {BatchNorm 2D};
            \node (relu1) [activation, below=of bn1] {ReLU};
            \node (z0) [embedding, below=of relu1, yshift=-1*\unit] {$\bm{z}_0$};
            \node (rearrange-1) [rearrange, below=of z0] {Rearrange\\from: (b, d, h, w)\\\ \ \ \ to: (b, h, w, d)};
            \node (tm1) [tokenmixer, below=of rearrange-1, yshift=-1.5*\unit] {TokenMixer};
            \node (cm1) [channelmixer, below=of tm1] {ChannelMixer};
            \node (zdg) [embedding, below=of cm1, yshift=-1*\unit] {$\bm{z}_{D_G}$};
            \node (ln2) [layernorm, below=of zdg] {LayerNorm};
            \node (avg1) [layer, below=of ln2] {Average(h, w)};
            \node (z-out) [embedding, below=of avg1, yshift=-1*\unit] {$\widehat{\bm{z}}$};

            \node (ar-split) [emptylayer, minimum width=0*\unit, below=of z-out] {};
            \node (a-start) [emptylayer, left=of ar-split] {};
            \node (a-linear) [linear, below=of a-start] {Linear};
            \node (a-out) [embedding, below=of a-linear, yshift=-1*\unit] {$\widehat{\bm{y}}$};
            \node (r-start) [emptylayer, right=of ar-split] {};
            \node (r-linear-1) [linear, below=of r-start] {Linear};
            \node (r-gelu) [activation, below=of r-linear-1] {GELU};
            \node (r-linear-2) [linear, below=of r-gelu] {Linear};
            \node (r-out) [embedding, below=of r-linear-2, yshift=-1*\unit] {$\widehat{\bm{r}}$};
        \end{scope}

        \draw (rbg) -- (con1) -- (bn1) -- (relu1);
        \draw [arrow] (relu1) -- (z0);
        \draw (z0) -- (rearrange-1) -- (tm1) -- (cm1);
        \draw [arrow] (cm1) -- (zdg);
        \draw (zdg) -- (ln2) -- (avg1);
        \draw [arrow] (avg1) -- (z-out);

        \draw [rounded corners=5pt] (z-out) -- (ar-split.center) -- (a-start.center) -- (a-linear);
        \draw [arrow] (a-linear) -- (a-out);
        \draw [rounded corners=5pt] (z-out) -- (ar-split.center) -- (r-start.center) -- (r-linear-1) -- (r-gelu) -- (r-linear-2);
        \draw [arrow] (r-linear-2) -- (r-out);

        \begin{scope}[on background layer]
            \node (bg_local) [fit={(con1.north west) (relu1.south east)}, background, fill=black!5, inner sep=0.5*\unit] {};
        \end{scope}
        \node (localx) [anchor=west, align=center, yshift=0.5*\unit, inner sep=0] at (bg_local.north west) {$\times D_L$};

        \begin{scope}[on background layer]
            \node (bg_global) [fit={(tm1) (cm1)}, background, fill=black!5, inner sep=0.5*\unit] {};
        \end{scope}
        \node (globalx) [anchor=west, align=center, yshift=0.5*\unit, inner sep=0] at (bg_global.north west) {$\times D_G$};

        \begin{scope}[node distance=0.6*\unit]
            \node (tm-z-input) [embedding, right=of rbg.north east, anchor=north, xshift=13*\unit] {$\bm{z}_{l-1}$};
            \node (tm-main-res-start) [emptylayer, minimum height=0*\unit, below=of tm-z-input, yshift=-0.6*\unit] {};
            \node (tm-ln-1) [layernorm, below=of tm-main-res-start] {LayerNorm};
            \node (tm-sp-res-start) [emptylayer, minimum height=0*\unit, below=of tm-ln-1] {};
            \node (tm-rearrange-1) [rearrange, below=of tm-sp-res-start] {Rearrange\\from: (b, h, w, d)\\\ \ \ \ to: (b, d, h, w)};
            \node (tmconv1) [conv2d, below=of tm-rearrange-1] {Conv2D};
            \node (tm-rearrange-2) [rearrange, below=of tmconv1] {Rearrange\\from: (b, d, h, w)\\\ \ \ \ to: (b, h, w, d)};
            \node (tm-resplus-1) [resplus, below=of tm-rearrange-2] {$+$};

            \node (tm-path-split) [emptylayer, below=of tm-resplus-1] {};

            \node (tm-path-h-start) [emptylayer, left=of tm-path-split] {};
            \node (tm-path-h-rearrange-1) [rearrange, below=of tm-path-h-start] {Rearrange\\from: (b, h, w, (e * s))\\\ \ \ \ to: (b, w, e, (h * s))};
            \node (tm-path-h-linear) [linear, below=of tm-path-h-rearrange-1] {Linear};
            \node (tm-path-h-rearrange-2) [rearrange, below=of tm-path-h-linear] {Rearrange\\from: (b, w, e, (h * s))\\\ \ \ \ to: (b, h, w, (e * s))};
            \node (tm-path-h-end) [emptylayer, below=of tm-path-h-rearrange-2] {};

            \node (tm-path-w-start) [emptylayer, right=of tm-path-split] {};
            \node (tm-path-w-rearrange-1) [rearrange, below=of tm-path-w-start] {Rearrange\\from: (b, h, w, (e * s))\\\ \ \ \ to: (b, h, e, (w * s))};
            \node (tm-path-w-linear) [linear, below=of tm-path-w-rearrange-1] {Linear};
            \node (tm-path-w-rearrange-2) [rearrange, below=of tm-path-w-linear] {Rearrange\\from: (b, h, e, (w * s))\\\ \ \ \ to: (b, h, w, (e * s))};
            \node (tm-path-w-end) [emptylayer, below=of tm-path-w-rearrange-2] {};

            \node (tm-path-d-linear) [linear] at ($ (tm-path-h-linear)!0.5!(tm-path-w-linear) $) {Linear};

            \node (tm-concat-start) [emptylayer] at ($ (tm-path-h-end)!0.5!(tm-path-w-end) $) {};
            \node (tm-concat) [layer, below=of tm-concat-start] {Concatenate};
            \node (tm-linear) [linear, below=of tm-concat] {Linear};
            \node (tm-main-res-end) [resplus, below=of tm-linear] {$+$};
            \node (tm-z-out) [embedding, below=of tm-main-res-end, yshift=-0.5*\unit] {$\bm{z}^*_l$};
        \end{scope}

        \draw (tm-z-input) -- (tm-ln-1) -- (tm-rearrange-1) -- (tmconv1) -- (tm-rearrange-2) -- (tm-resplus-1) -- (tm-path-d-linear) -- (tm-concat) -- (tm-linear) -- (tm-main-res-end);
        \draw [arrow] (tm-main-res-end) -- (tm-z-out);

        \node (tmres2) at ($ (tm-rearrange-1.east |- tm-sp-res-start.center) + (0.5*\unit, 0) $) {};
        \node (tmres3) at ($ (tm-rearrange-1.east |- tm-resplus-1.center) + (0.5*\unit, 0) $) {};
        \draw [rounded corners=5pt] (tm-sp-res-start.center) -- (tmres2.center) -- (tmres3.center) -- (tm-resplus-1.east);

        \draw [rounded corners=5pt] (tm-path-split.center) -- (tm-path-h-start.center) -- (tm-path-h-rearrange-1) -- (tm-path-h-linear) -- (tm-path-h-rearrange-2) -- (tm-path-h-end.center) -- (tm-concat-start.center);
        \draw [rounded corners=5pt] (tm-path-split.center) -- (tm-path-w-start.center) -- (tm-path-w-rearrange-1) -- (tm-path-w-linear) -- (tm-path-w-rearrange-2) -- (tm-path-w-end.center) -- (tm-concat-start.center);

        \node (tm-main-res-ne) at ($ (tm-path-w-rearrange-1.east |- tm-main-res-start.center) + (0.5*\unit, 0) $) {};
        \node (tm-main-res-se) at ($ (tm-path-w-rearrange-1.east |- tm-main-res-end.center) + (0.5*\unit, 0) $) {};
        \draw [rounded corners=5pt] (tm-main-res-start.center) -- (tm-main-res-ne.center) -- (tm-main-res-se.center) -- (tm-main-res-end.east);

        \begin{scope}[on background layer]
            \node (bg-tm) [fit={(tm-main-res-start) (tm-path-h-rearrange-1) (tm-main-res-ne) (tm-main-res-end)}, background, fill=tokenmixer!10, inner sep=0.5*\unit] {};
        \end{scope}
        \node (bg-tm-label) [anchor=west, align=center, yshift=0.5*\unit, inner sep=0] at (bg-tm.north west) {Token Mixer};

        \begin{scope}[node distance=0.6*\unit]
            \node (cm-res-start) [emptylayer, minimum height=0*\unit, below=of tm-z-out, yshift=-0.6*\unit] {};
            \node (cm-ln-1) [layernorm, below=of cm-res-start] {LayerNorm};
            \node (cm-linear-1) [linear, below=of cm-ln-1] {Linear};
            \node (cm-gelu-1) [activation, below=of cm-linear-1] {GELU};
            \node (cm-linear-2) [linear, below=of cm-gelu-1] {Linear};
            \node (cm-res-end) [resplus, below=of cm-linear-2] {$+$};
            \node (cm-z-out) [embedding, below=of cm-res-end, yshift=-0.5*\unit] {$\bm{z}_l$};
        \end{scope}

        \draw (tm-z-out) -- (cm-ln-1) -- (cm-linear-1) -- (cm-gelu-1) -- (cm-linear-2) -- (cm-res-end);
        \draw [arrow] (cm-res-end) -- (cm-z-out);

        \node (cm-res-ne) at ($ (cm-ln-1.east |- cm-res-start.center) + (0.5*\unit, 0) $) {};
        \node (cm-res-se) at ($ (cm-ln-1.east |- cm-res-end.center) + (0.5*\unit, 0) $) {};
        \draw [rounded corners=5pt] (cm-res-start.center) -- (cm-res-ne.center) -- (cm-res-se.center) -- (cm-res-end.east);

        \begin{scope}[on background layer]
            \node (bg-cm) [fit={(cm-res-start) (cm-ln-1) (cm-res-ne) (cm-res-end)}, background, fill=channelmixer!10, inner sep=0.5*\unit] {};
        \end{scope}
        \node (bg-cm-label) [anchor=west, align=left, xshift=-1*\unit, yshift=1*\unit, inner sep=0] at (bg-cm.north west) {Channel\\Mixer};

    \end{tikzpicture}
    \caption{
    \textbf{UMAVR architecture.}
    The left side demonstrates UMAVR's processing of the unified input matrix $\chi^{\text{RPM}}$ resulting in the predicted index of the correct answer $\widehat{y}$, as well as the predicted rule representation $\widehat{r}$.
    The right side illustrates the architecutre of TokenMixer and ChannelMixer modules.
    The einops notation~\cite{rogozhnikov2022einops} is used to denote tensor transformations in the Rearrange module, where b, h, w, d denote the batch, height, width, and feature dimension, resp.
    }
    \label{fig:model}
\end{figure}

\section{UMAVR architecture}\label{sec:umavr-architecture}
Figure~\ref{fig:model} illustrates the architecture of UMAVR.

\begin{figure}[t]
    \centering
    \includegraphics[width=.98\textwidth]{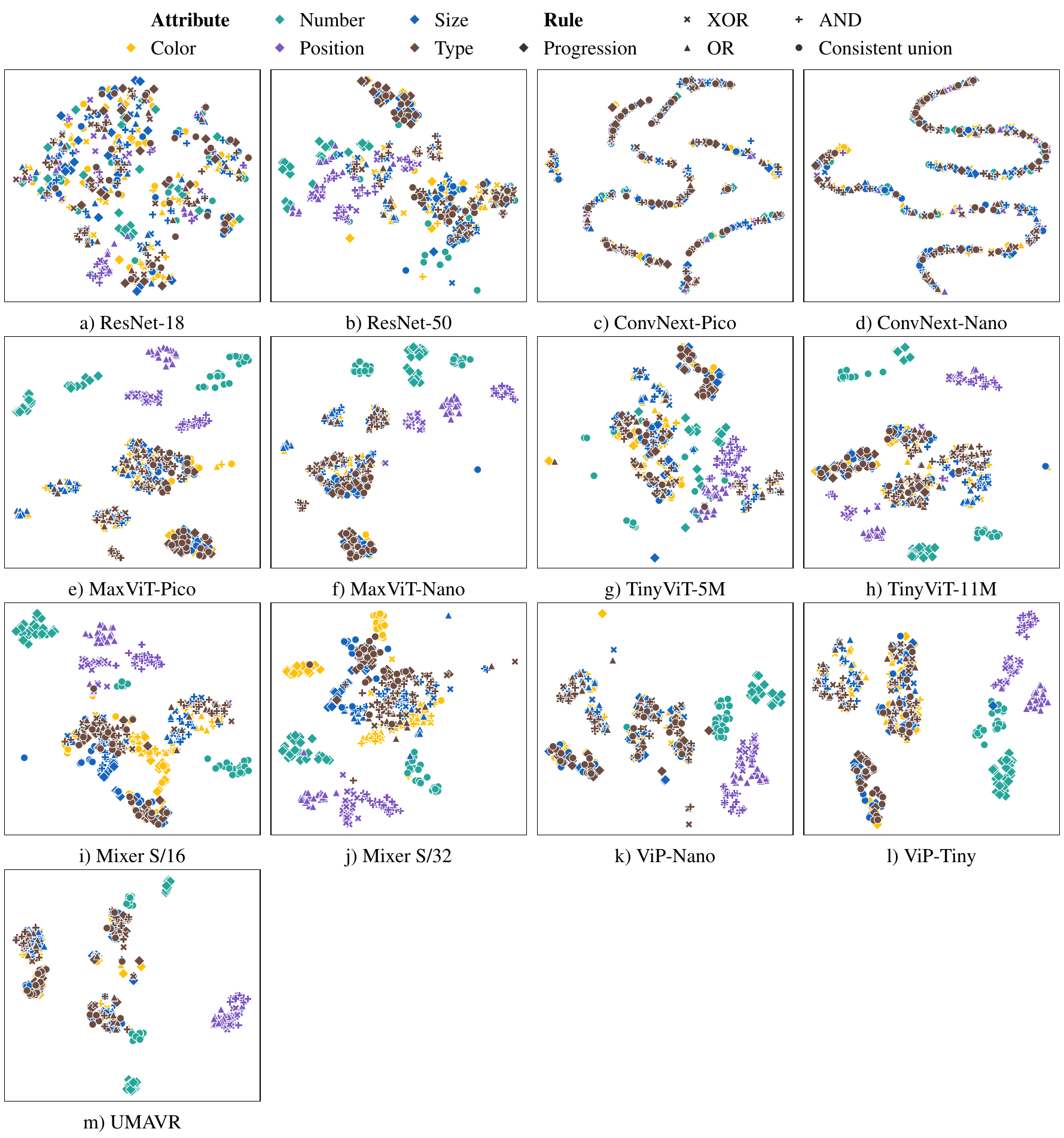}
    \caption{
    \textbf{PGM embeddings.}
    The embeddings of PGM matrices ($n_a=2$) from the test split of the \texttt{Neutral} regime, visualized with t-SNE~\cite{van2008visualizing}.
    For the sake of interpretability, the figure considers matrices with a single rule applied to Shape objects.}
    \label{fig:embeddings-pgm-4x4}
\end{figure}

\begin{figure}[t]
    \centering
    \includegraphics[width=.98\textwidth]{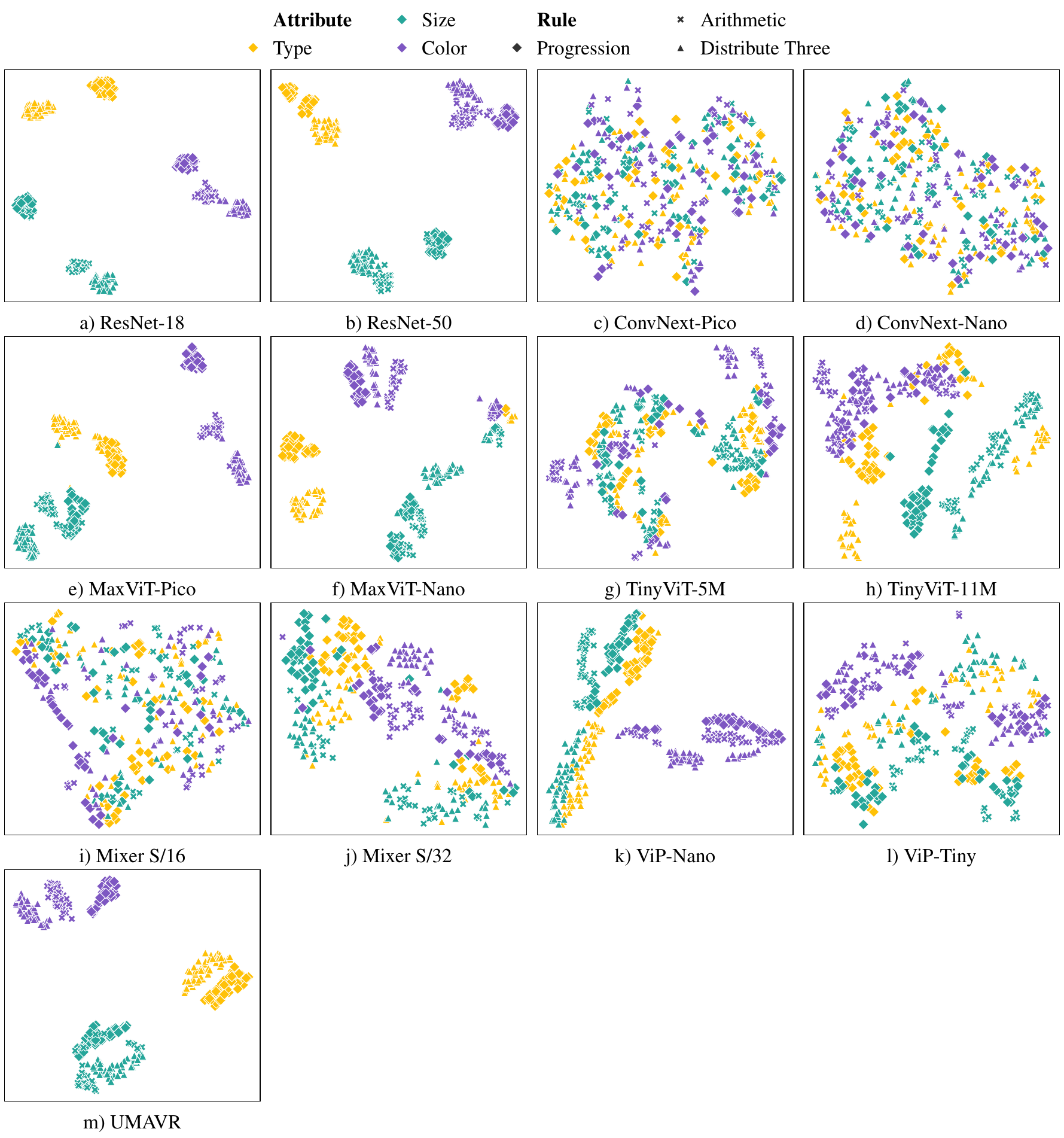}
    \caption{
    \textbf{I-RAVEN embeddings.}
    The embeddings of I-RAVEN matrices ($n_a=2$) from the test split of the \texttt{Center-Single} configuration, visualized with t-SNE~\cite{van2008visualizing}.
    For the sake of interpretability we consider matrices in which all but one attributes are governed by the Constant rule.
    }
    \label{fig:embeddings-raven-4x4}
\end{figure}

\begin{figure}[t]
    \centering
    \includegraphics[width=.98\textwidth]{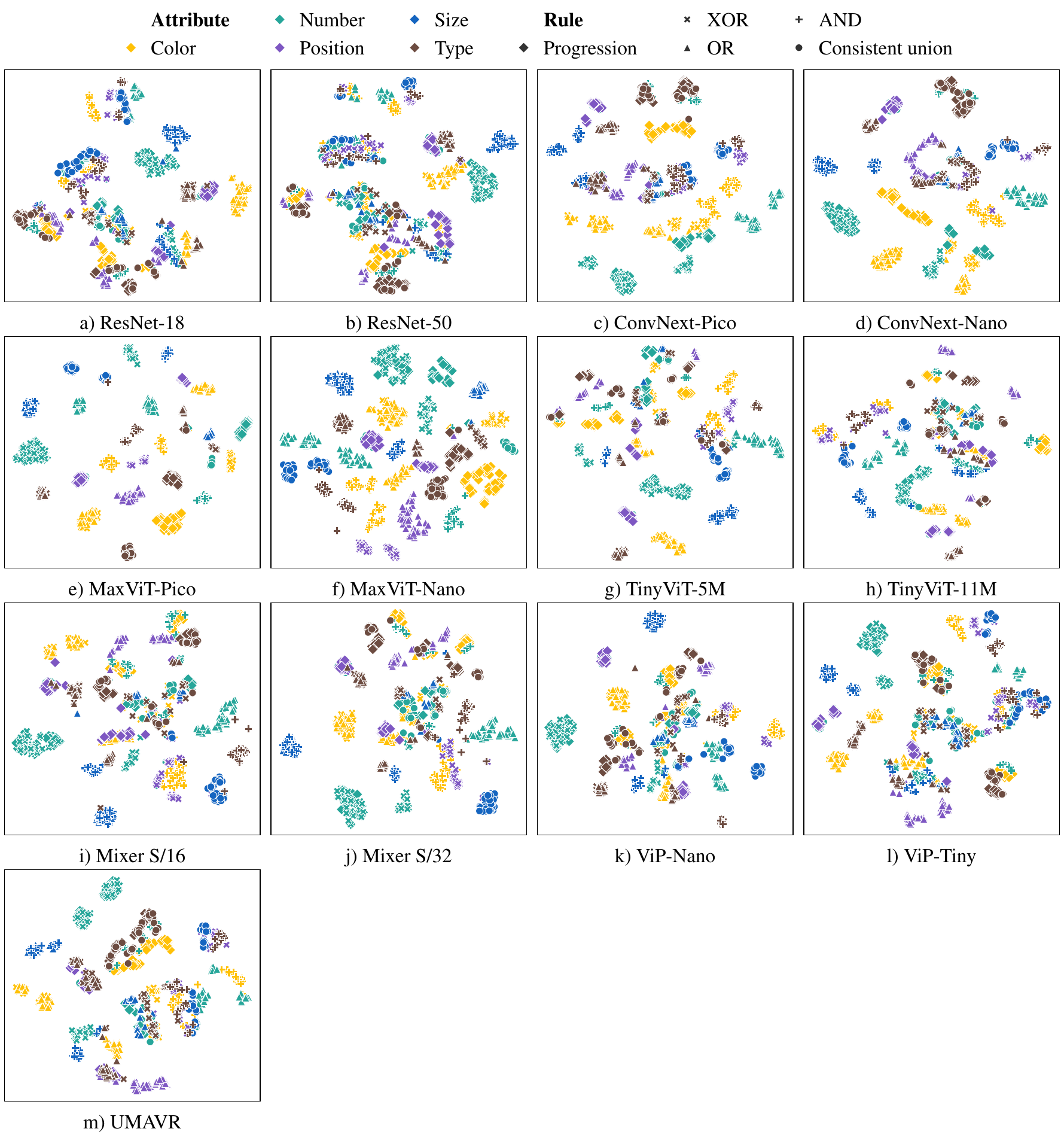}
    \caption{
    \textbf{VAP embeddings.}
    The embeddings of VAP matrices ($n_a=2$) from the test split, visualized with t-SNE~\cite{van2008visualizing}.
    }
    \label{fig:embeddings-vap-4x4}
\end{figure}

\section{Comparison to disjoint representation}\label{sec:app:comparison}
For the most demanding setting of $n_a=n_a^{\text{max}}$, the best-performing unified approaches outperformed the state-of-the-art disjoint result on G-set ($97.6\%$ vs. $82.8\%$~\cite{tomaszewska2022duel}), were on-par on PGM ($97.3\%$ vs. $98.2\%$~\cite{mondal2023learning}) and VAP ($98.4\%$ vs. $98.5\%$~\cite{mondal2023learning}), and were inferior on I-RAVEN ($86.9\%$ vs. $95.7\%$~\cite{mondal2023learning}) and VASR ($62.3\%$ vs. $86.0\%$\cite{bitton2023vasr}). 

While the comparison shows a slight advantage of disjoint representation, it is important to note that the unified view opens several research avenues and poses challenges that extend beyond the sole performance improvement.
Specifically, the use of the unified representation allows bridging the domains of broad CV and AVR by means of enabling the development of methods that could be seamlessly applied to both areas.
Furthermore, the unified representation allows employing the current CV methods operating on single images to solve AVR tasks, and to use pre-trained checkpoints for their initialization.
Finally, as mentioned in Section~\ref{sec:conclusion}, universal methods developed for solving uniformly viewed AVR tasks may accelerate progress in other domains that require relational reasoning, via knowledge reuse.

\section{Embedding visualization}\label{sec:embedding-visualization}

\paragraph{PGM.}
Figure~\ref{fig:embeddings-pgm-4x4} extends Figure~\ref{fig:embeddings} from the main paper and compares PGM matrix embeddings ($n_a=2$) across all models listed in Table~\ref{tab:single-task-learning}.
These plots lead to the same conclusion -- the clustering quality correlates with model's performance on the target task, which shows that the models indeed learned to identify (at least to some extent) the underlying abstract rules instead of relying on visual shortcuts.
Embeddings of ConvNext models, which achieved results at the random guess level, don't present any distinct clusters.
In remaining models, we generally observe clusters of matrices with rules applied to Number and Position attributes (green and purple, resp.).
Embeddings of leading models in this setting, both Mixer variants, additionally display clusters concerning the Color attribute (yellow).
Nevertheless, Size and Type attributes (blue and purple, resp.), remain difficult to disentangle for all considered models.
Correct identification of these attributes may require a relatively large receptive field, in comparison to Color, Number and Position which can be recognized even from low-level features.

\paragraph{I-RAVEN.}
Figure~\ref{fig:embeddings-raven-4x4} compares UMAVR matrix embeddings ($n_a=2$) across all models on I-RAVEN.
In general, the clustering quality correlates with model's performance on the target task (cf. Table~\ref{tab:single-task-learning}), e.g., the sequence of clusters produced with TinyViT-5M ($83.5\%$), ViP-Nano ($88.3\%$) and UMAVR ($95.6\%$) presents better and better quality.
However, the embeddings of both ResNet variants and MaxViT-Pico, which performed at the random guess level, form distinct clusters as well.
As described in the experimental setting, the models are trained with two loss functions: the main one evaluates model's accuracy in selecting the correct answer, while the auxiliary one assesses model's performance in identifying the underlying rules.
These models learned to identify the matrix rules almost perfectly, as presented in their embedding visualization.
At the same time, they failed to apply the rules to select the correct answer to the matrix, as demonstrated by their results on the target task ($\sim50.0\%$).
The task of identifying matrix rules is fundamentally simpler -- the correct rule can be deduced by comparing the 1st and 2nd row of the matrix.
However, correct application of the rule involves analysing the 3rd row of the matrix, as well as choosing between the provided answer panels, which explains the discrepancy in models' performance.
We conducted ablation experiments in which both ResNets and MaxViT-Pico were trained without the auxiliary loss function to solve I-RAVEN matrices.
However, similarly to the base setting, the models were unable to perform better than random guessing, which suggests that I-RAVEN poses a challenge for these models even in the simplest setting with $n_a=2$.

\paragraph{VAP.}
Figure~\ref{fig:embeddings-vap-4x4} visualizes embeddings of all models evaluated on VAP.
In comparison to Figures~\ref{fig:embeddings-raven-4x4} and~\ref{fig:embeddings-pgm-4x4}, all evaluated models provide clusterings of comparable quality, which aligns with their high STL results presented in Table~\ref{tab:single-task-learning}.
The results echo analogous conclusions to those from the main paper, namely that the models learned to identify the underlying abstract rules in VAP matrices, and do not rely on visual shortcuts or dataset biases to solve the task.

\begin{figure}[t]
    \centering
    \begin{subfigure}[t]{0.3\textwidth}
        \includegraphics[width=\textwidth]{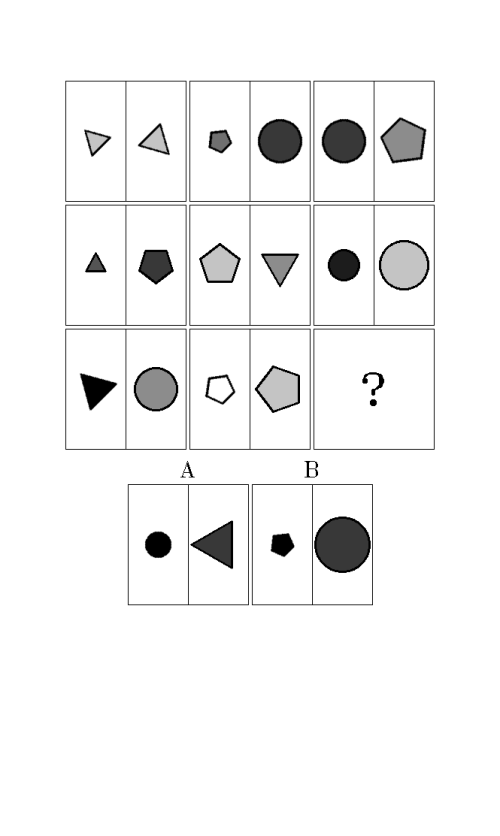}
        \caption{$n_a=2$}
        \label{fig:rpm-iraven-1}
    \end{subfigure}
    \hfill
    \begin{subfigure}[t]{0.3\textwidth}
        \includegraphics[width=\textwidth]{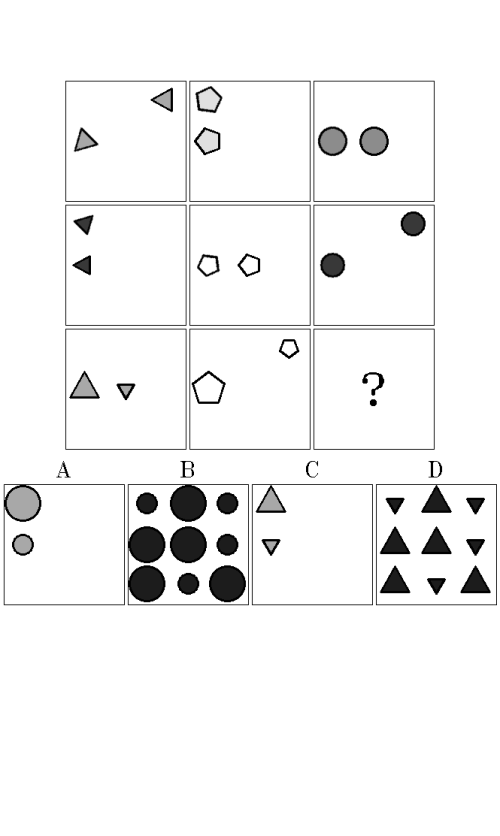}
        \caption{$n_a=4$}
        \label{fig:rpm-iraven-2}
    \end{subfigure}
    \hfill
    \begin{subfigure}[t]{0.3\textwidth}
        \includegraphics[width=\textwidth]{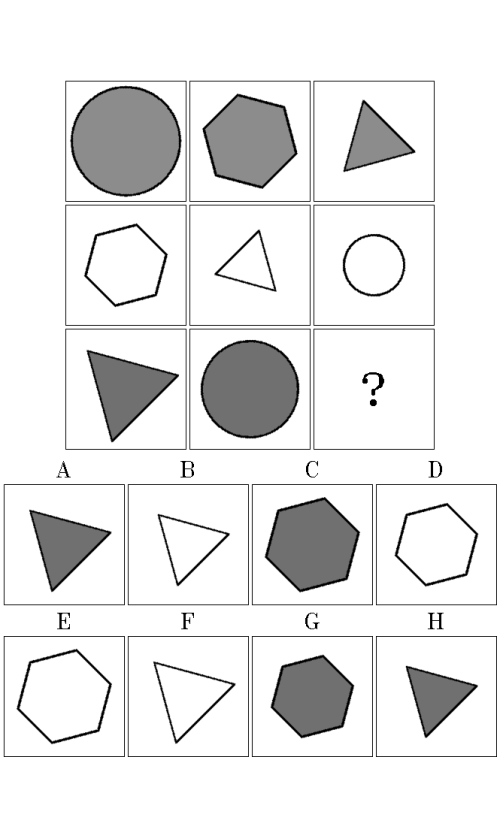}
        \caption{$n_a=8$}
        \label{fig:rpm-iraven-3}
    \end{subfigure}
    \caption{\textbf{RPMs from I-RAVEN~\cite{zhang2019raven,hu2021stratified}.}
    The appropriate answer panel (A-B, A-D, A-H, resp.) has to be chosen to complete the $3 \times 3$ context grid.
    The examples belong to the three selected I-RAVEN configurations: \texttt{Left-Right}, \texttt{3x3}, and \texttt{Center}, respectively.
    Each problem instance has up to 8 rules, which can be applied separately to each matrix hierarchy (e.g., only to left parts of each panel in (a)).
    The matrices are governed by the following rules:
    (a) in each row there is exactly one panel with the left part presenting a triangle, a pentagon, and a circle, respectively;
    (b) in each row there is exactly one panel containing triangles, pentagons, and circles, resp., and each image contains exactly two shapes;
    (c) in each row there is exactly one panel presenting a circle, a hexagon, and a triangle, resp., the shapes in a given row are of the same colour, and their size decreases from left to right.
    The correct answers are A, A and G, respectively.
    }
    \label{fig:rpm-iraven}
\end{figure}

\begin{figure}[t]
    \centering
    \begin{subfigure}[t]{0.3\textwidth}
        \includegraphics[width=\textwidth]{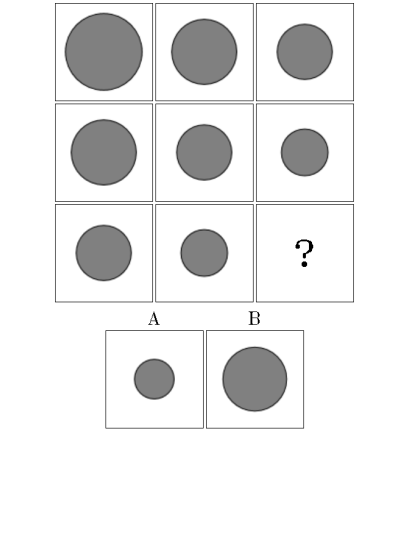}
        \caption{$n_a=2$}
        \label{fig:rpm-deepiq-1}
    \end{subfigure}
    \hfill
    \begin{subfigure}[t]{0.3\textwidth}
        \includegraphics[width=\textwidth]{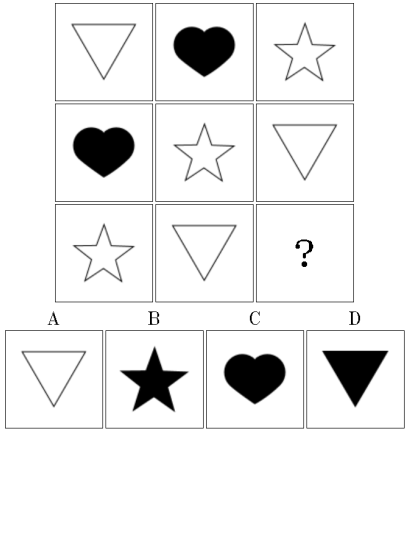}
        \caption{$n_a=4$}
        \label{fig:rpm-deepiq-2}
    \end{subfigure}
    \hfill
    \begin{subfigure}[t]{0.3\textwidth}
        \includegraphics[width=\textwidth]{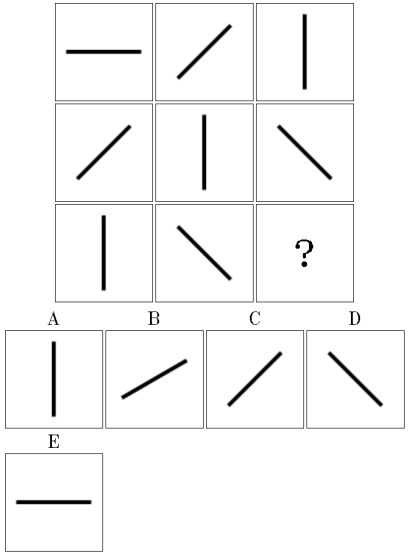}
        \caption{$n_a=5$}
        \label{fig:rpm-deepiq-3}
    \end{subfigure}
    \caption{\textbf{RPMs from G-set~\cite{mandziuk2019deepiq,tomaszewska2022duel}.}
    The appropriate answer panel (A-B, A-D, A-E, resp.) has to be chosen to complete the $3 \times 3$ context grid.
    The matrices are governed by the following rules:
    (a) progression applied to object size;
    (b) each row has the same three object shapes;
    (c) progression applied to object rotation.
    The correct answers are A, C and E, respectively.}
    \label{fig:rpm-gset}
\end{figure}

\begin{figure}[t]
    \centering
    \begin{subfigure}[t]{0.3\textwidth}
        \includegraphics[width=\textwidth]{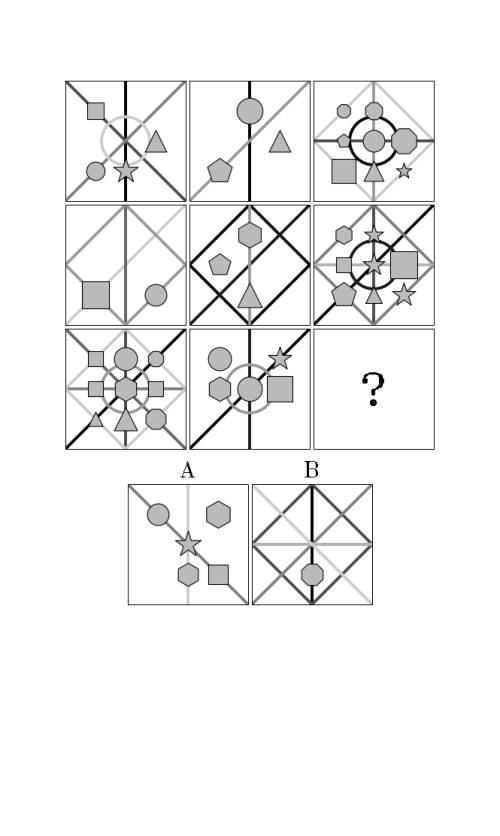}
        \caption{$n_a=2$}
        \label{fig:rpm-pgm-1}
    \end{subfigure}
    \hfill
    \begin{subfigure}[t]{0.3\textwidth}
        \includegraphics[width=\textwidth]{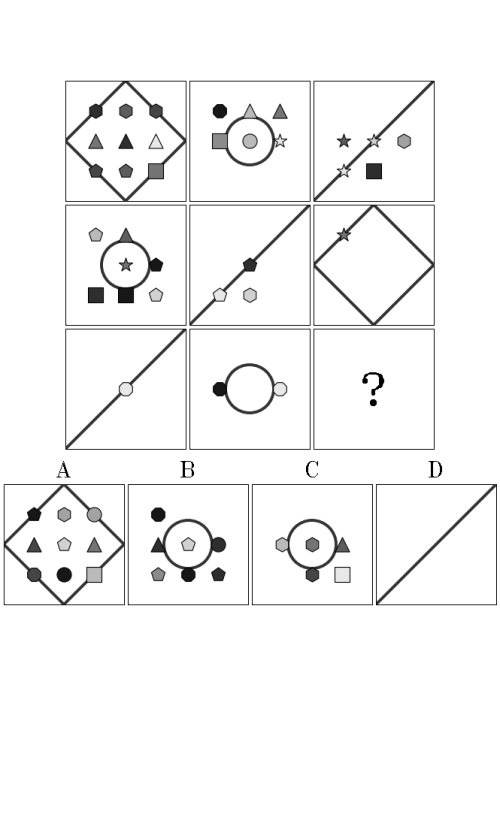}
        \caption{$n_a=4$}
        \label{fig:rpm-pgm-2}
    \end{subfigure}
    \hfill
    \begin{subfigure}[t]{0.3\textwidth}
        \includegraphics[width=\textwidth]{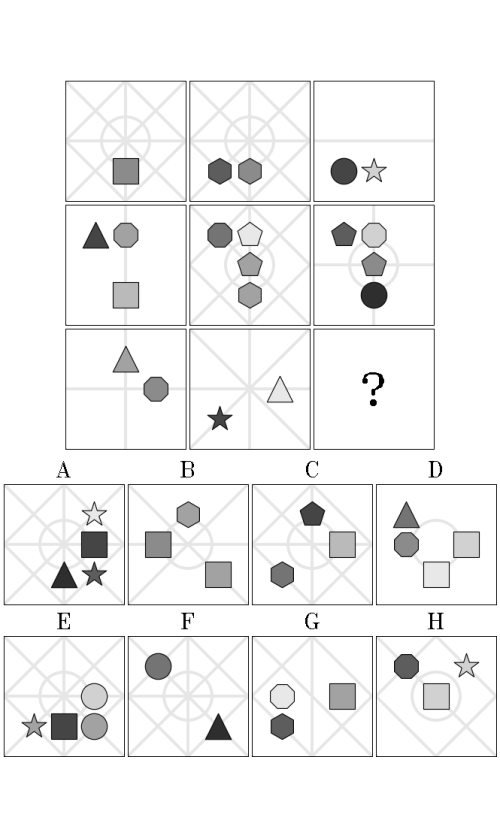}
        \caption{$n_a=8$}
        \label{fig:rpm-pgm-3}
    \end{subfigure}
    \caption{\textbf{RPMs from PGM~\cite{santoro2018measuring}.}
    The appropriate answer panel (A-B, A-D, A-H, resp.) has to be chosen to complete the $3 \times 3$ context grid.
    The samples belong to the Neutral PGM regime.
    The matrices are governed by the following rules:
    (a) \texttt{OR} applied row-wise to line colour;
    (b) \texttt{consistent union} applied row-wise to line type;
    (c) \texttt{OR} applied row-wise to shape position.
    The correct answers are B, A and C, respectively.
    The examples additionally demonstrate the role of distracting features in the PGM matrices (e.g. shapes other than lines in example (a)).
    These features don't conform to any rule, as their goal is to increase the difficulty of the task.}
    \label{fig:rpm-pgm}
\end{figure}

\begin{figure}[t]
    \centering
    \begin{subfigure}[t]{0.4\textwidth}
        \includegraphics[width=\textwidth]{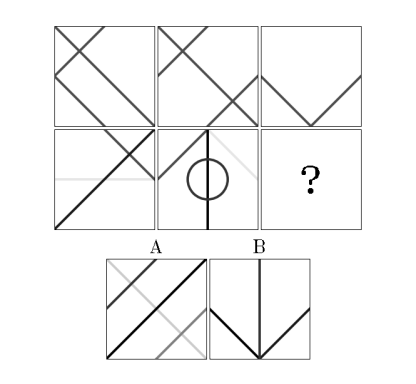}
        \caption{$n_a=2$}
        \label{fig:vap-1}
    \end{subfigure}
    \hfil
    \begin{subfigure}[t]{0.4\textwidth}
        \includegraphics[width=\textwidth]{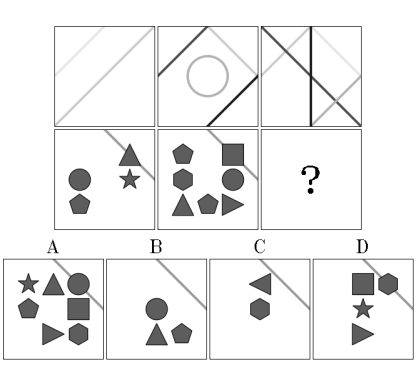}
        \caption{$n_a=4$}
        \label{fig:vap-2}
    \end{subfigure}
    \caption{\textbf{VAPs~\cite{hill2018learning}.}
    The appropriate answer panel (A-B, A-D, resp.) has to be chosen to complete the $2 \times 3$ context grid.
    The samples come from the Novel Domain Transfer regime, in which a rule demonstrated in a source domain (first row) has to be abstracted and applied to a novel target domain (second row).
    The matrices are governed by the following rules:
    (a) \texttt{XOR} abstracted from line type to line colour;
    (b) \texttt{OR} abstracted from line colour to shape type.
    The correct answers are B and A, respectively.}
    \label{fig:vap}
\end{figure}

\begin{figure}[t]
    \centering
    \begin{subfigure}[t]{0.49\textwidth}
        \includegraphics[width=\textwidth]{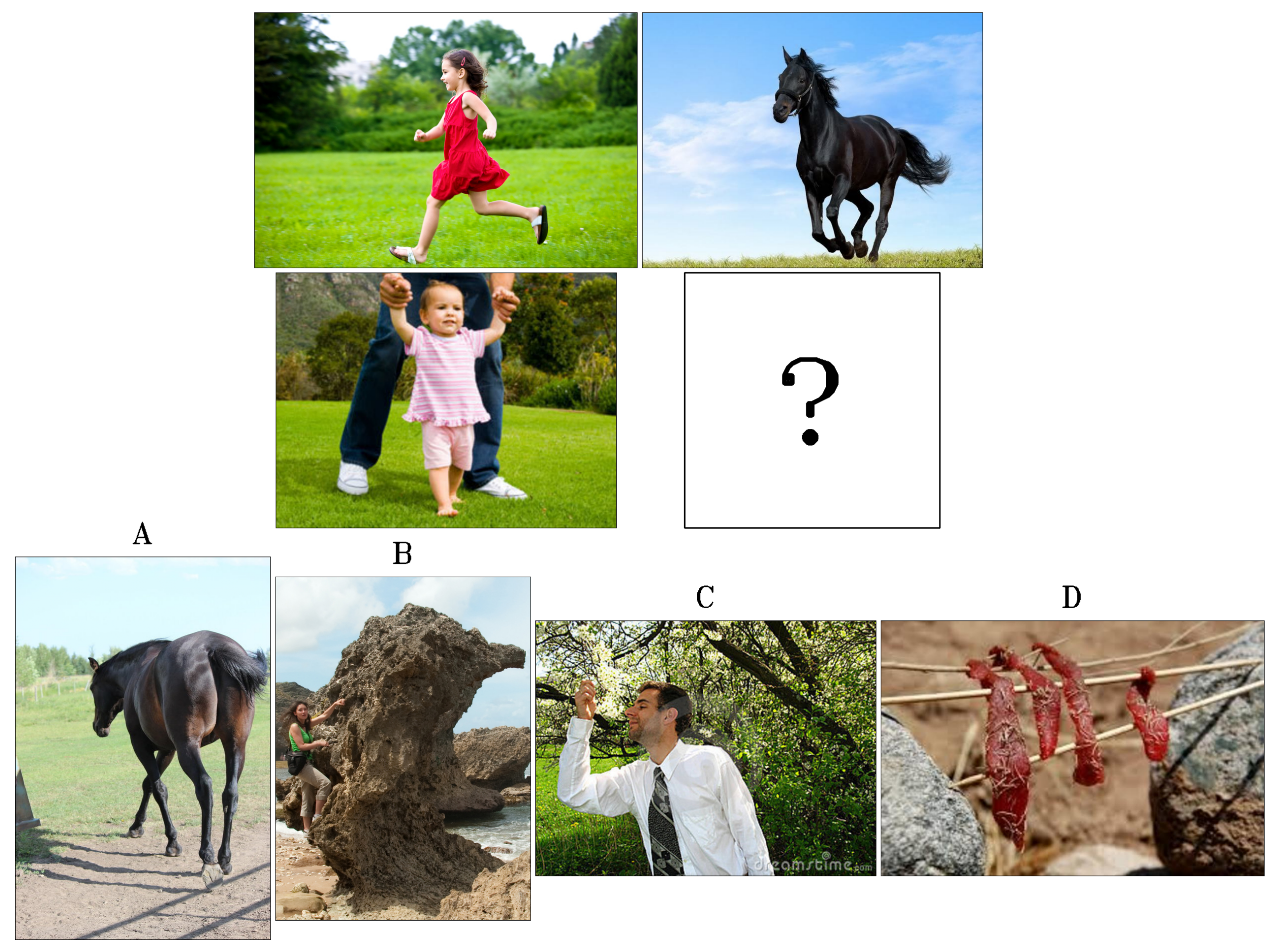}
        \caption{}
        \label{fig:vasr-1}
    \end{subfigure}
    \hfil
    \begin{subfigure}[t]{0.49\textwidth}
        \includegraphics[width=\textwidth]{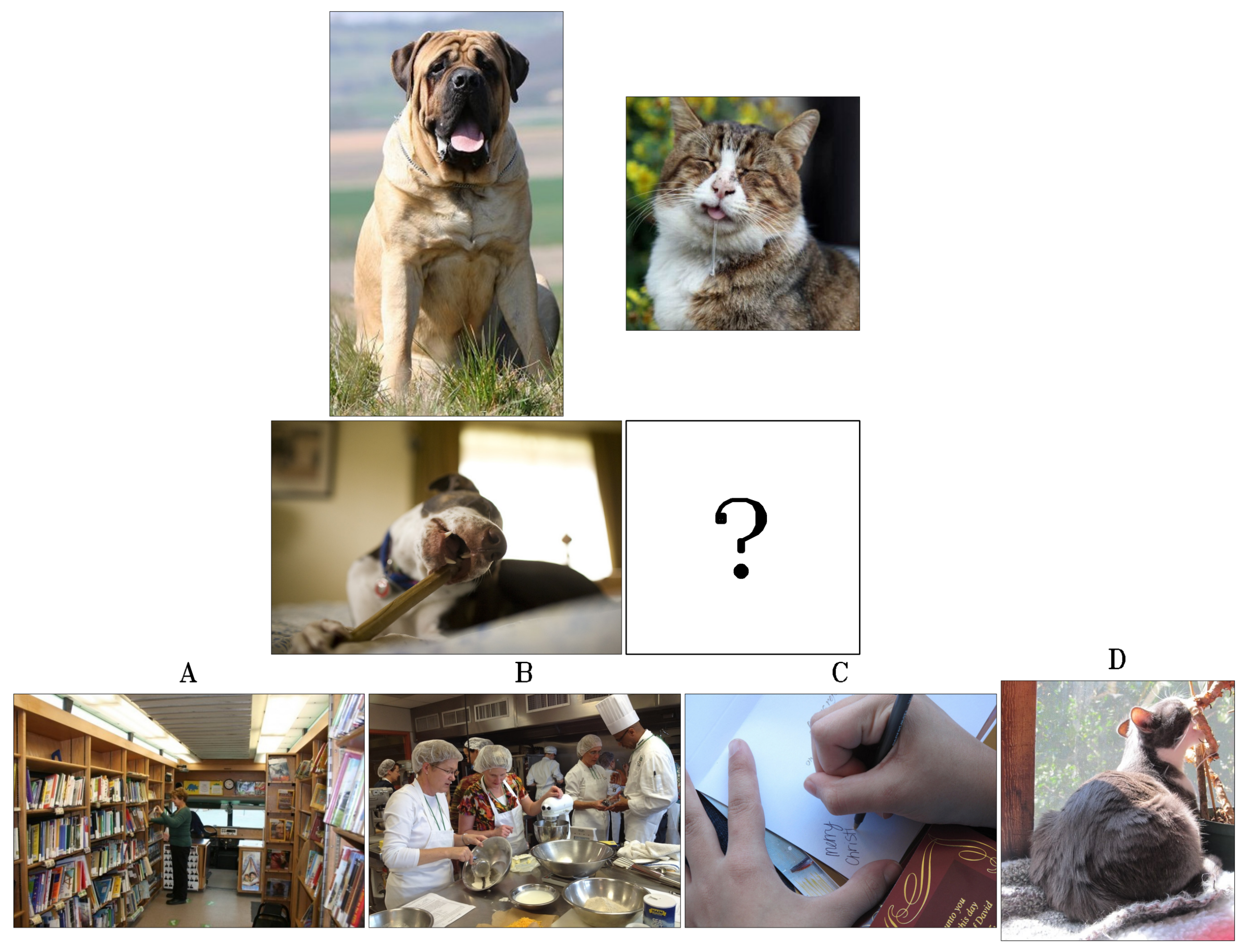}
        \caption{}
        \label{fig:vasr-2}
    \end{subfigure}
    \caption{\textbf{VASR matrices~\cite{bitton2023vasr}.}
    The appropriate answer panel (A -- D) has to be chosen to complete the $2 \times 2$ context grid.
    The samples were taken from the dataset variant with random distractors.
    The correct answers are A and D, respectively.}
    \label{fig:vasr}
\end{figure}

\end{document}